\newif\ifdraft\draftfalse

\documentclass[review=false,manuscript]{acmart}

\acmJournal{PACMPL}
\acmVolume{1}
\acmNumber{CONF} 
\acmArticle{1}
\acmYear{2018}
\acmMonth{1}
\acmDOI{} 
\startPage{1}

\setcopyright{none}

\usepackage{booktabs}   
\usepackage{subcaption} 

\newcommand{\ottdrule}[4][]{{\displaystyle\frac{\begin{array}{l}#2\end{array}}{#3}\quad\ottdrulename{#4}}}

\newcommand{\ottpremise}[1]{ #1 \\}
\newenvironment{ottdefnblock}[3][]{ \framebox{\mbox{#2}} \quad #3 \\[0pt]}{}

\newcommand{\ottnt}[1]{\mathit{#1}}
\newcommand{\ottmv}[1]{\mathit{#1}}

\newcommand{\ottsym}[1]{#1}

\newcommand{\ottdrulename}[1]{\textsc{#1}}

\newcommand{\ottdruleHole}[1]{\ottdrule[#1]{%
}{
\Gamma  \vdash  \left[ \, \right] \,  \mathrel{:}  \, \ottnt{P}}{%
{\ottdrulename{Hole}}{}%
}}

\newcommand{\ottdruleVar}[1]{\ottdrule[#1]{%
\ottpremise{\mathit{x}  \mathord{:}  \ottnt{P} \,  \in  \, \Gamma}%
}{
\Gamma  \vdash  \mathit{x} \,  \mathrel{:}  \, \ottnt{P}}{%
{\ottdrulename{Var}}{}%
}}

\newcommand{\ottdruleAbs}[1]{\ottdrule[#1]{%
\ottpremise{\Gamma  \ottsym{,}  \mathit{x}  \mathord{:}  \ottnt{P}  \vdash  \ottnt{M} \,  \mathrel{:}  \, \ottnt{Q}}%
}{
\Gamma  \vdash  \lambda  \mathit{x}  \ottsym{.}  \ottnt{M} \,  \mathrel{:}  \, \ottnt{P} \,  \mathord{\rightarrow}  \, \ottnt{Q}}{%
{\ottdrulename{Abs}}{}%
}}

\newcommand{\ottdruleApp}[1]{\ottdrule[#1]{%
\ottpremise{ \Gamma  \vdash  \ottnt{M} \,  \mathrel{:}  \, \ottnt{P} \,  \mathord{\rightarrow}  \, \ottnt{Q}  \quad  \Gamma  \vdash  \ottnt{N} \,  \mathrel{:}  \, \ottnt{P} }%
}{
\Gamma  \vdash  \ottnt{M} \, \ottnt{N} \,  \mathrel{:}  \, \ottnt{Q}}{%
{\ottdrulename{App}}{}%
}}

\newcommand{\ottdrulePair}[1]{\ottdrule[#1]{%
\ottpremise{ \Gamma  \vdash  \ottnt{M} \,  \mathrel{:}  \, \ottnt{P}  \quad  \Gamma  \vdash  \ottnt{N} \,  \mathrel{:}  \, \ottnt{Q} }%
}{
\Gamma  \vdash  \ottsym{(}  \ottnt{M}  \ottsym{,}  \ottnt{N}  \ottsym{)} \,  \mathrel{:}  \, \ottnt{P} \,  \mathord{\times}  \, \ottnt{Q}}{%
{\ottdrulename{Pair}}{}%
}}

\newcommand{\ottdruleCasePair}[1]{\ottdrule[#1]{%
\ottpremise{ \Gamma  \vdash  \ottnt{M} \,  \mathrel{:}  \, \ottnt{P} \,  \mathord{\times}  \, \ottnt{Q}  \quad  \Gamma  \ottsym{,}  \mathit{x}  \mathord{:}  \ottnt{P}  \ottsym{,}  \mathit{y}  \mathord{:}  \ottnt{Q}  \vdash  \ottnt{N} \,  \mathrel{:}  \, \ottnt{R} }%
}{
\Gamma  \vdash  \mathsf{case} \, \ottnt{M} \, \mathsf{of} \, \ottsym{(}  \mathit{x}  \ottsym{,}  \mathit{y}  \ottsym{)}  \rightarrow  \ottnt{N} \,  \mathrel{:}  \, \ottnt{R}}{%
{\ottdrulename{CasePair}}{}%
}}

\newcommand{\ottdruleLeft}[1]{\ottdrule[#1]{%
\ottpremise{\Gamma  \vdash  \ottnt{M} \,  \mathrel{:}  \, \ottnt{P}}%
}{
\Gamma  \vdash  \mathsf{Left} \, \ottnt{M} \,  \mathrel{:}  \, \ottnt{P} \,  \mathord{+}  \, \ottnt{Q}}{%
{\ottdrulename{Left}}{}%
}}

\newcommand{\ottdruleRight}[1]{\ottdrule[#1]{%
\ottpremise{\Gamma  \vdash  \ottnt{M} \,  \mathrel{:}  \, \ottnt{Q}}%
}{
\Gamma  \vdash  \mathsf{Right} \, \ottnt{M} \,  \mathrel{:}  \, \ottnt{P} \,  \mathord{+}  \, \ottnt{Q}}{%
{\ottdrulename{Right}}{}%
}}

\newcommand{\ottdruleCaseSum}[1]{\ottdrule[#1]{%
\ottpremise{ \Gamma  \vdash  \ottnt{L} \,  \mathrel{:}  \, \ottnt{P} \,  \mathord{+}  \, \ottnt{Q}  \quad   \Gamma  \ottsym{,}  \mathit{x}  \mathord{:}  \ottnt{P}  \vdash  \ottnt{M} \,  \mathrel{:}  \, \ottnt{R}  \quad  \Gamma  \ottsym{,}  \mathit{y}  \mathord{:}  \ottnt{Q}  \vdash  \ottnt{N} \,  \mathrel{:}  \, \ottnt{R}  }%
}{
\Gamma  \vdash  \mathsf{case} \, \ottnt{L} \, \mathsf{of} \, \ottsym{\{} \, \mathsf{Left} \, \mathit{x}  \rightarrow  \ottnt{M}  \ottsym{;} \, \mathsf{Right} \, \mathit{y}  \rightarrow  \ottnt{N}  \ottsym{\}} \,  \mathrel{:}  \, \ottnt{R}}{%
{\ottdrulename{CaseSum}}{}%
}}

\usepackage{algorithm,algorithmicx,algpseudocode}
\usepackage{amsmath}
\usepackage{ebproof}
\usepackage{multirow}
\usepackage{tabu}

\newcommand{\TS}[1]{\ifdraft\textcolor{red}{#1}\fi}
\newcommand{\KS}[1]{\ifdraft\textcolor{blue}{#1}\fi}

\newcommand{\sect}[1]{Section~\ref{sec:#1}}
\newcommand{\fig}[1]{Figure~\ref{fig:#1}}
\newcommand{\defn}[1]{Definition~\ref{defn:#1}}
\newcommand{\tbl}[1]{Table~\ref{tbl:#1}}
\newcommand{\proc}[1]{Procedure~\ref{alg:#1}}

\floatname{algorithm}{Procedure}

\DeclareMathOperator*{\argmax}{arg\,max}

\newcommand{\Real}{\ensuremath{\mathbb{R}}}
\newcommand{\RealMat}[2]{\Real^{{#1} \mathbin{\times} {#2}}}
\newcommand{\RealVec}[1]{\Real^{{#1}}}

\newcommand{\Wparam}[2]{\ensuremath{W_{#2}^\mathrm{#1}}}
\newcommand{\bparam}[2]{\ensuremath{b_{#2}^\mathrm{#1}}}
\newcommand{\Fact}[1]{\ensuremath{F^\mathrm{#1}}}

\newcommand{\queue}{\ensuremath{\mathcal{Q}}}
\newcommand{\prob}{\ensuremath{\mathcal{P}}}

\newcommand{\dataset}{\textrm{D}}
\newcommand{\datasetall}{\ensuremath{\mathrm{D}_\mathbf{all}}}
\newcommand{\datasett}{\ensuremath{\mathrm{D}_\mathbf{t}}}
\newcommand{\datasetv}{\ensuremath{\mathrm{D}_\mathbf{v}}}
\newcommand{\datasetsmall}{\ensuremath{\mathrm{D}_\mathbf{small}}}
\newcommand{\datasetlarge}{\ensuremath{\mathrm{D}_\mathbf{large}}}
\newcommand{\termset}{\textrm{S}}

\newcommand{\probvar}[1]{\ensuremath{\boldsymbol{{#1}}}}

\newcommand{\CTXS}{\ensuremath{\mathbf{Ctx}}}

\newcommand{\invin}{\rotatebox[origin=c]{180}{\ensuremath{\in}}}

\newcommand{\TYPEOF}{\textsf{typeof}}

\begin{document}

\title[]{Automated proof synthesis for propositional logic with deep neural networks}


\author{Taro Sekiyama}
\affiliation{
  \institution{National Institute of Informatics}
  \country{Japan}
}
\email{sekiyama@nii.ac.jp}          

\author{Kohei Suenaga}
\affiliation{
  \institution{Kyoto University}
  \country{Japan}
}
\email{ksuenaga@kuis.kyoto-u.ac.jp}         
\affiliation{
  \institution{JST PRESTO}
  \country{Japan}
}
\email{ksuenaga@kuis.kyoto-u.ac.jp}         


\begin{abstract}
  This work explores the application of \emph{deep learning}, a
  machine learning technique that uses \emph{deep neural networks
    (DNN)} in its core, to an automated theorem proving (ATP) problem.
  %
  %
  %
  To this end, we construct a statistical model which quantifies the likelihood
  that a proof is indeed a correct one of a given proposition.
  Based on this model, we give a proof-synthesis procedure that
  searches for a proof in the order of the likelihood.
  This procedure uses an estimator of the likelihood of an inference
  rule being applied at each step of a proof.
  As an implementation of the estimator, we propose a
  \emph{proposition-to-proof} architecture, which is a DNN
  tailored to the automated proof synthesis problem.
  %
  %
  To empirically demonstrate its usefulness, we apply our model to
  synthesize proofs of propositional logic.
  We train the proposition-to-proof model using a training dataset of
  proposition--proof pairs.
  The evaluation against a benchmark set shows the very high accuracy and an
  improvement to the recent work of neural proof synthesis.
\end{abstract}



\maketitle

\ifdraft
\section{TODO: REMOVE THIS BEFORE SUBMITTING THE PAPER}
\begin{itemize}
 \item dummy
\end{itemize}
\fi

\section{Introduction}



\emph{Theorem proving} is an essential activity in formal reasoning.
Needless to say, mathematics has become the reliable foundation of
modern natural science, including several branches of theoretical
computer science, by justifying theorems with proofs.
The importance of correct proofs leads to the study of software called
proof
assistants~\cite{Nipkow:2002:IPA:1791547,CoqManualV8,Norell:2009:DTP:1481861.1481862},
which allow users to state theorems and their proofs formally in the
form of certain programming languages and automatically check that the
proofs correctly prove the theorems.
The realm of the areas that rely on theorem proving is expanding
beyond mathematics; for example, it is being applied for system
verification~\cite{Klein:2009:SFV:1629575.1629596,Leroy:2009:FVR:1538788.1538814},
where one states the correctness of a system as a theorem and
justifies it in the form of proofs.

\emph{Automated theorem proving
  (ATP)}~\cite{bibel2013automated,pfenning2004automated,fitting2012first}
is a set of techniques that prove logical formulas automatically.
We are concerned with the following form of ATP called \emph{automated
  proof synthesis (APS)}: Given a logical formula $\ottnt{P}$, if $\ottnt{P}$
holds, return a proof $\ottnt{M}$ of $\ottnt{P}$.
In the light of the importance of theorem proving, APS serves as a
useful tool for activities based on formal reasoning.
For example, from the perspective of the aforementioned system
verification, APS serves for automating system verification; indeed,
various methods for (semi)automated static program
verification~\cite{DBLP:conf/vstte/BarnettDFJLSV05,Chalin:2007:IVE:1292316.1292322,DBLP:conf/esop/FilliatreP13}
can be seen as APS procedures.
We also remark another important application of APS: automated program
synthesis.  An APS algorithm can be seen as an automated program
synthesis procedure via the Curry--Howard
isomorphism~\cite{Sorensen:2006:LCI:1197021}, in which $\ottnt{M}$ can be
seen as a program and $\ottnt{P}$ can be seen as a
specification.
Not only is APS interesting from the practical viewpoint, it is also
interesting from the theoretical perspective in that it investigates
the algorithmic aspect of theorem proving.

Traditionally, the main weapon from the programming-language community
to tackle APS has been \emph{symbolic} methods; an APS algorithm
inspects the syntactic structure of the formula $\ottnt{P}$ and, using the
obtained information, tries to construct a proof derivation of
$\ottnt{P}$.
A seminal work in this regard is by \citet{benyelles79};
they proposed a sound and complete APS algorithm for an implicational
fragment of the propositional logic.
%

This paper tackles the APS problem using another emerging technology:
\emph{statistical machine learning}.
In particular, we explore an application of \emph{deep neural networks
  (DNN)}~\cite{Goodfellow-et-al-2016}.  DNNs have seen a great success
in recent years for solving various tasks; to name a few, image
recognition~\cite{he2016deep}, speech
recognition~\cite{hinton2012deep}, and natural language
processing~\cite{bahdanau2014neural,cho2014properties,wu2016google}.
%
%
To this end, we propose a novel DNN architecture named
\emph{proposition-to-proof} model\footnote{In the community of neural
  network research, there is a habit to call a trained DNN ``model''.
  Following this convention, we abuse the word ``model'' for a trained
  DNN.} tailored to the APS problem.

Concretely, we statistically model the APS problem in terms of
probabilities.  This statistical model serves for quantifying how a
partially constructed proof is likely to lead to a correct proof of
the given proposition $\ottnt{P}$.
Based on this statistical model, we define a proof-synthesis procedure
that searches for a proof of given proposition $\ottnt{P}$ in the order
of the likelihood.
This proof synthesis procedure requires a function to estimate the
likelihood of an inference rule being applied at a specific step of
a proof (or, equivalently, a specific position of a partially
constructed proof).  For this estimation, we use a DNN based on
the proposition-to-proof architecture that we propose.
We empirically evaluate the performance of our network, which reveals that it
can predict the inference rules that fill the rest of a partially constructed
proof of a propositional-logic formula with $96.79\%$ accuracy.

This work is not the first one that applies DNNs to APS.  Among them,
\citet{DBLP:journals/corr/SekiyamaIS17} reports an
application of DNN equipped with long-short term memory (LSTM) to the
APS problem.
The idea in their work is to view the APS problem as a machine
translation problem from the language of logical formulas to the
language of proofs.
Based on this view, they applied an off-the-shelf neural machine
translation framework to APS.
They report that their network, which is a vanilla one for neural
machine translation, proved around $50\%$ of the propositional-logic
formulas in the benchmark they used.
%
In contrast to their approach of trying to synthesize an entire proof at once,
we designed our proof-synthesis procedure so that it \emph{gradually} constructs
a proof based on the likelihood of each inference rule.

The contributions of this work are summarized as follows.
\begin{itemize}
\item We construct a statistical model for the APS problem in terms of
      probability.
  This model formally quantifies the likelihood of a proof being a correct one
      of a given proposition.  Applying the laws of probability to this
  model, we derive how the computation of the likelihood of an entire
  proof is reduced to the successive computations of the
  likelihood of inference rules.
\item Based on this statistical model, we design a proof-synthesis
  procedure that searches for a proof of a given proposition in the
  descending order of the likelihood.  This algorithm gradually
  constructs a proof by repeating the estimation of the
  likelihood of occurrences of inference rules in the proof.
\item We propose a novel DNN architecture which we call
  \emph{proposition-to-proof} model that estimates the above
  likelihood of inference rules.  This network takes a proposition
  $\ottnt{P}$, the position in a partially constructed proof to be
  filled, and contextual information as input and outputs the
  likelihood of inference rules being applied at the
  position to be filled.
\item We implemented the proof-synthesis procedure with a trained
  proposition-to-proof model and empirically confirmed its
  effectiveness compared to \citet{DBLP:journals/corr/SekiyamaIS17}.  In addition to
  measuring the accuracy of the trained proposition-to-proof model, we
  conducted in-depth analyses of the model.  We confirmed that our
  model estimates the proof constructor with $96.79\%$ accuracy.


\end{itemize}

Currently, we do not claim that our procedure outperforms the
state-of-the-art APS method for propositional logic.
Rather, our contribution consists in the statistical reformulation of
the APS problem and application of deep learning, which exposes superhuman
performance in many areas.
We believe that deep learning is also useful in the APS problem possibly in
combination with symbolic methods and that the present work opens up a
new research direction in this regard.

The rest of this paper is organized as follows:
\sect{stlc} defines the logic and the proof system that we
use in this paper;
\sect{statisticalModeling} reviews statistical machine learning briefly;
\sect{proofConstruction} defines the proof-synthesis
algorithm;
\sect{model-deep} gives a brief introduction to deep learning and introduces the
proposition-to-proof architecture;
\sect{exp} describes the result of the experiments;
\sect{relatedWork} discusses related work; and
\sect{conc} concludes.

We assume the readers' familiarity to the Curry--Howard
isomorphism~\cite{Sorensen:2006:LCI:1197021}.
We sometimes abuse the terminologies in the simply typed
lambda-calculus for those of the proof theory of the propositional
logic.
We also assume that the readers are familiar with the probability
theory although we do not deal with measure-theoretic discussions in
the present paper.

\section{The Simply Typed Lambda Calculus as Propositional Logic}
\label{sec:stlc}
In this work, we identify the simply typed lambda calculus with
the intuitionistic propositional logic via the Curry--Howard
isomorphism~\cite{Sorensen:2006:LCI:1197021}.
This view is indeed beneficial for us:
%
(1) a term of the simply typed lambda calculus is the concise
representation of a derivation tree, which is essentially the proof of
a proposition and
(2) we can express a partially constructed proof as a term with
\emph{holes}, which denote positions in a proof that needs to be
filled.
%
%
%
In the rest of this section, we introduce the simply typed lambda
calculus extended with product types and sum types.
The Curry--Howard isomorphism allows us to identify a product type
with the conjunction of propositions and a sum type with the
disjunction.
%

\begin{figure}[t]
 \[\begin{array}{rcl}
  \multicolumn{3}{l}{\textbf{Types}} \\[.5ex]
    \ottnt{P}, \ottnt{Q}, \ottnt{R} &::=&
     \mathit{a} \mid \ottnt{P} \,  \mathord{\rightarrow}  \, \ottnt{Q} \mid \ottnt{P} \,  \mathord{\times}  \, \ottnt{Q} \mid \ottnt{P} \,  \mathord{+}  \, \ottnt{Q}
  \\[1ex]

  \multicolumn{3}{l}{\textbf{Terms}} \\[.5ex]
    \ottnt{L}, \ottnt{M}, \ottnt{N} &::=&
      \left[ \, \right]  \mid \mathit{x} \mid
     \lambda  \mathit{x}  \ottsym{.}  \ottnt{M} \mid \ottnt{M} \, \ottnt{N} \mid
     \ottsym{(}  \ottnt{M}  \ottsym{,}  \ottnt{N}  \ottsym{)} \mid \mathsf{case} \, \ottnt{M} \, \mathsf{of} \, \ottsym{(}  \mathit{x}  \ottsym{,}  \mathit{y}  \ottsym{)}  \rightarrow  \ottnt{N} \mid \\ &&
     \mathsf{Left} \, \ottnt{M} \mid \mathsf{Right} \, \ottnt{M} \mid
     \mathsf{case} \, \ottnt{L} \, \mathsf{of} \, \ottsym{\{} \, \mathsf{Left} \, \mathit{x}  \rightarrow  \ottnt{M}  \ottsym{;} \, \mathsf{Right} \, \mathit{y}  \rightarrow  \ottnt{N}  \ottsym{\}}
  \\[1ex]

  \multicolumn{3}{l}{\textbf{Typing contexts}} \\[.5ex]
    \Gamma &::=&  \emptyset  \mid \Gamma  \ottsym{,}  \mathit{x}  \mathord{:}  \ottnt{P}
  \\
   \end{array}\]
 \caption{Syntax.}
 \label{fig:syntax}
\end{figure}

\begin{figure}[t]
 \begin{flushleft}
  \qquad \framebox{$\Gamma  \vdash  \ottnt{M} \,  \mathrel{:}  \, \ottnt{P}$} \\[1ex]
 \end{flushleft}
 \begin{center}
  $\ottdruleHole{}$ \hfil
  $\ottdruleVar{}$ \\[1ex]
  $\ottdruleAbs{}$ \hfil
  $\ottdruleApp{}$ \\[1ex]
  $\ottdrulePair{}$ \hfil
  $\ottdruleCasePair{}$ \\[1ex]
  $\ottdruleLeft{}$ \hfil
  $\ottdruleRight{}$ \\[1ex]
  $\ottdruleCaseSum{}$ \hfil
 \end{center}
 \caption{Inference rules.}
 \label{fig:infrule}
\end{figure}

\fig{syntax} shows the syntax of the simply typed lambda calculus.
\emph{Types} (or \emph{propositions}) are represented by the metavariables $\ottnt{P}$, $\ottnt{Q}$,
and $\ottnt{R}$; \emph{terms} (or \emph{proofs}) are represented by the metavariables
$\ottnt{L}$, $\ottnt{M}$, and $\ottnt{N}$; and \emph{typing contexts} (or \emph{collections of assumptions}) are
represented by the metavariable $\Gamma$.
%
%
The definition of types is standard: they consist of \emph{type
  variables} (or \emph{propositional variables}), \emph{function types} $\ottnt{P} \,  \mathord{\rightarrow}  \, \ottnt{Q}$, \emph{product
  types} $\ottnt{P} \,  \mathord{\times}  \, \ottnt{Q}$, and \emph{sum types} $\ottnt{P} \,  \mathord{+}  \, \ottnt{Q}$.
We use the metavariables $\mathit{a}$, $\mathit{b}$, $\mathit{c}$, and $\mathit{d}$ for
type variables.
The syntax of \emph{terms} is that of the simply typed lambda calculus.
Products are constructed by $\ottsym{(}  \ottnt{M}  \ottsym{,}  \ottnt{N}  \ottsym{)}$ and destructed by
$\mathsf{case} \, \ottnt{M} \, \mathsf{of} \, \ottsym{(}  \mathit{x}  \ottsym{,}  \mathit{y}  \ottsym{)}  \rightarrow  \ottnt{M}$; sums are constructed by
$\mathsf{Left} \, \ottnt{M}$ and $\mathsf{Right} \, \ottnt{N}$ and destructed by
$\mathsf{case} \, \ottnt{L} \, \mathsf{of} \, \ottsym{\{} \, \mathsf{Left} \, \mathit{x}  \rightarrow  \ottnt{M}  \ottsym{;} \, \mathsf{Right} \, \mathit{y}  \rightarrow  \ottnt{N}  \ottsym{\}}$.
The term syntax is equipped with a \emph{hole} $ \left[ \, \right] $ to express
partially constructed terms.  A hole denotes a position in a term that
needs to be filled (see below).
We use metavariables $\mathit{x}$, $\mathit{y}$, and $\mathit{z}$ for term variables.

The notions of free variables, bound variables,
and substitution for terms are defined as usual.
$\lambda  \mathit{x}  \ottsym{.}  \ottnt{M}$ binds $\mathit{x}$ in $\ottnt{M}$; $\mathsf{case} \, \ottnt{M} \, \mathsf{of} \, \ottsym{(}  \mathit{x}  \ottsym{,}  \mathit{y}  \ottsym{)}  \rightarrow  \ottnt{N}$ binds
$\mathit{x}$ and $\mathit{y}$ in $\ottnt{N}$; and
$\mathsf{case} \, \ottnt{L} \, \mathsf{of} \, \ottsym{\{} \, \mathsf{Left} \, \mathit{x}  \rightarrow  \ottnt{M}  \ottsym{;} \, \mathsf{Right} \, \mathit{y}  \rightarrow  \ottnt{N}  \ottsym{\}}$ binds $\mathit{x}$ in $\ottnt{M}$ and
$\mathit{y}$ in $\ottnt{N}$, respectively.
Types have no binders.
We write $\textsf{FV} \, \ottsym{(}  \ottnt{M}  \ottsym{)}$ for the set of term variables that occur freely
in $\ottnt{M}$.
We write $ [  \ottnt{N}  /  \mathit{x}  ]  \, \ottnt{M}$ for the capture-avoiding substitution of $\ottnt{N}$
for $\mathit{x}$ in $\ottnt{M}$.
We say two terms are $\alpha$-equivalent if they are different only in
the use of bound variable names.
We identify two $\alpha$-equivalent terms.
%
%
%

A term that contains holes represents a partially constructed proof.
Our proof-synthesis procedure introduced in
\sect{proofConstruction} maintains a set of partially constructed
terms and fills a hole inside a term in the set at each step.
%
%
We assume that holes in a term are uniquely identified by natural
numbers.  We write $ \left[ \, \right] _{ \ottmv{i} } $ for a hole with number $i$.
%
%
%
We write $ \ottnt{M}  [  \ottnt{N}  ]_{ \ottmv{i} } $ for the term obtained by filling the hole
$ \left[ \, \right] _{ \ottmv{i} } $ in $\ottnt{M}$ with $\ottnt{N}$.

We also define the typing relation $\Gamma  \vdash  \ottnt{M} \,  \mathrel{:}  \, \ottnt{P}$ as the least
relation that satisfies the inference rules in \fig{infrule}.
This relation means that term $\ottnt{M}$ is typed at $\ottnt{P}$ under
$\Gamma$ or, equivalently, $\ottnt{M}$ is a proof of $\ottnt{P}$ under assumptions $\Gamma$.
We write $\Gamma  \not\vdash  \ottnt{M} \,  \mathrel{:}  \, \ottnt{P}$ to denote that $\Gamma  \vdash  \ottnt{M} \,  \mathrel{:}  \, \ottnt{P}$ does not
hold.
%
%
The rules in \fig{infrule} are standard except for the rule
$\ottdrulename{Hole}$ for holes.
This rule allows any type to be given to a hole.
We call the inference rules except for the rule $\ottdrulename{Hole}$
\emph{proof inference rules}.
We say that $\ottnt{M}$ is a \emph{(complete) proof} of $\ottnt{P}$ if
$ \emptyset   \vdash  \ottnt{M} \,  \mathrel{:}  \, \ottnt{P}$ is derived and $\ottnt{M}$ has no holes.
$\ottnt{M}$ is said to be \emph{partial} or \emph{partially constructed} if
$ \emptyset   \vdash  \ottnt{M} \,  \mathrel{:}  \, \ottnt{P}$ but $\ottnt{M}$ contains holes.

\begin{figure}[t]
 \begin{flushleft}
  \qquad \framebox{$\ottnt{M} \,  \longrightarrow_\beta  \, \ottnt{N}$} \quad \textbf{$\beta$-reduction} \\[1ex]
 \end{flushleft}
 \[\begin{array}{rcl}
  \ottsym{(}  \lambda  \mathit{x}  \ottsym{.}  \ottnt{M}  \ottsym{)} \, \ottnt{N} & \longrightarrow_\beta &  [  \ottnt{N}  /  \mathit{x}  ]  \, \ottnt{M} \\[1ex]
  \mathsf{case} \, \ottsym{(}  \ottnt{L}  \ottsym{,}  \ottnt{M}  \ottsym{)} \, \mathsf{of} \, \ottsym{(}  \mathit{x}  \ottsym{,}  \mathit{y}  \ottsym{)}  \rightarrow  \ottnt{N} & \longrightarrow_\beta &  [  \ottnt{L}  /  \mathit{x}  ,  \ottnt{M}  /  \mathit{y}  ]  \, \ottnt{N} \\[1ex]
  \mathsf{case} \, \ottsym{(}  \mathsf{Left} \, \ottnt{L}  \ottsym{)} \, \mathsf{of} \, \ottsym{\{} \, \mathsf{Left} \, \mathit{x}  \rightarrow  \ottnt{M}  \ottsym{;} \, \mathsf{Right} \, \mathit{y}  \rightarrow  \ottnt{N}  \ottsym{\}} & \longrightarrow_\beta &  [  \ottnt{L}  /  \mathit{x}  ]  \, \ottnt{M} \\[1ex]
  \mathsf{case} \, \ottsym{(}  \mathsf{Right} \, \ottnt{L}  \ottsym{)} \, \mathsf{of} \, \ottsym{\{} \, \mathsf{Left} \, \mathit{x}  \rightarrow  \ottnt{M}  \ottsym{;} \, \mathsf{Right} \, \mathit{y}  \rightarrow  \ottnt{N}  \ottsym{\}} & \longrightarrow_\beta &  [  \ottnt{L}  /  \mathit{y}  ]  \, \ottnt{N}
   \end{array}\]
 \\[2ex]
 \begin{flushleft}
  \qquad \framebox{$\ottnt{M} \,  \longrightarrow_\eta  \, \ottnt{N}$} \quad \textbf{$\eta$-reduction} \\[1ex]
 \end{flushleft}
 \[\begin{array}{rcll}
  \ottsym{(}  \lambda  \mathit{x}  \ottsym{.}  \ottnt{M} \, \mathit{x}  \ottsym{)} & \longrightarrow_\eta & \ottnt{M} & (\mathit{x} \,  \not\in  \, \textsf{FV} \, \ottsym{(}  \ottnt{M}  \ottsym{)}) \\
  \ottsym{(}  \mathsf{case} \, \ottnt{L} \, \mathsf{of} \, \ottsym{(}  \mathit{x}  \ottsym{,}  \mathit{y}  \ottsym{)}  \rightarrow  \mathit{x}  \ottsym{,}  \mathsf{case} \, \ottnt{L} \, \mathsf{of} \, \ottsym{(}  \mathit{x}  \ottsym{,}  \mathit{y}  \ottsym{)}  \rightarrow  \mathit{y}  \ottsym{)} & \longrightarrow_\eta & \ottnt{L} \\
  \multicolumn{4}{c}{
   \mathsf{case} \, \ottnt{M} \, \mathsf{of} \, \ottsym{\{} \, \mathsf{Left} \, \mathit{x}  \rightarrow   [  \mathsf{Left} \, \mathit{x}  /  \mathit{z}  ]  \, \ottnt{N}  \ottsym{;} \, \mathsf{Right} \, \mathit{y}  \rightarrow   [  \mathsf{Right} \, \mathit{y}  /  \mathit{z}  ]  \, \ottnt{N}  \ottsym{\}} \,  \longrightarrow_\eta  \,  [  \ottnt{M}  /  \mathit{z}  ]  \, \ottnt{N}
  } \\
  \multicolumn{4}{r}{(\mathit{x}, \mathit{y} \,  \not\in  \, \textsf{FV} \, \ottsym{(}  \ottnt{N}  \ottsym{)})}
   \end{array}\]
 \caption{Reduction rules.}
 \label{fig:redrule}
\end{figure}

To define the notion of normal forms, we introduce $\beta$-reduction
($ \longrightarrow_\beta $) and $\eta$-reduction ($ \longrightarrow_\eta $), which are the least
compatible relations satisfying the rules in \fig{redrule}; the last
$\eta$-reduction rule for sums is given by \citet{Ghani_1995_TLCA}.
Term $\ottnt{M}$ is a $\beta\eta$ normal form when there does not exist
$\ottnt{N}$ such that $\ottnt{M} \,  \longrightarrow_\beta  \, \ottnt{N}$ nor $\ottnt{M} \,  \longrightarrow_\eta  \, \ottnt{N}$.

\section{Background: statistical machine learning}
\label{sec:statisticalModeling}
\label{sec:statisticalMachineLearning}

In order to make the present paper self-contained, we explain basic
concepts on statistical machine learning and probabilities that appear
in this paper.
The exposition about machine learning in this
section is not intended to be exhaustive; for detail, see the standard
textbooks, e.g., \citet{Bishop_2006_PRML}.

\emph{Machine learning} is a generic term for a set of techniques to
make software ``learn'' how to behave from data without being
explicitly programmed.
The machine-learning task in this paper is of a type called
\emph{supervised learning}.  In this type of tasks, a learner needs to
synthesize a function $f : X \rightarrow Y$ for certain sets $X$ and
$Y$.\footnote{We only consider the case where $X$ and $Y$ are discrete
  in this paper.}
In a typical setting, the learner is given a set
$D := \{(x_1,y_1),\dots,(x_n,y_n)\}$ of sample input--output pairs of
$f$ as a hint for this learning task; the set $D$ is called a
\emph{training dataset}.
%

A popular strategy to tackle this problem is to use statistics.
In order to explain application of statistics in machine learning, we
fix notation about probabilities.
%
We designate a \emph{random variable}, say $\probvar{B}_x$, that
evaluates to an element of $Y$ following a certain probabilistic
distribution parameterized by an element $x \in X$; we use the bold
face for random variables in this paper.
For this random variable, one can consider, for example, the
probability $p(\probvar{B}_x=y)$ of $\probvar{B}_x$ being evaluated to
$y$.
More generally, given a predicate $\probvar{\varphi}(x,\probvar{B}_x)$
over $x$ and $\probvar{B}_x$, one can define the probability
$p(\probvar{\varphi})$ that $\probvar{\varphi}$ holds for $x$ and a
value of $\probvar{B}_x$.\footnote{Strictly speaking, we need to
  define the structure of measurable sets on $Y$ and argue that
  $\{y \in Y \mid \varphi(x,y)\}$ is measurable on this structure to
  formally define $p(\probvar{\varphi})$.  We do not discuss such
  measure-theoretic issues in this paper.}
Notice that the truth value of the predicate $\probvar{\varphi}$ is a
random variable; it may hold or may not hold depending on the result
of the evaluation of $\probvar{B}_x$ in general.
We can also define the probability
$p(\probvar{\varphi_1} \mid \probvar{\varphi_2})$ for given two
predicates $\probvar{\varphi_1}$ and $\probvar{\varphi_2}$, which is
the probability of \emph{$\probvar{\varphi_1}$ holding under the
  condition that $\probvar{\varphi_2}$ is true}.
A \emph{probability distribution} of a random variable $\probvar{B}_x$
conditioned on $\probvar{\varphi}$, written
$p(\probvar{B}_x \mid \probvar{\varphi})$, is a function that maps an
element $y \in Y$ to the probability
$p(\probvar{B}_x=y \mid \probvar{\varphi})$.
%

We view the statistical machine learning in the following way.
Ideally, it is desirable to discover the ``true'' probability distribution,
especially probability distribution $d_x$ behind the random variable
$\probvar{B}_x$, that explains how the training dataset $D$ is generated.
If we have this distribution, then we can guess a highly probable output $y$ to
$x$ as one that maximizes the value of $d_x$ (i.e., $\argmax_{y \in Y} d_x(y)$).
However, it is actually hard to identify $d_x$ precisely.
An alternative promising way is to approximate $d_x$ by a parameterized function
$F_{x,w_1,\dots,w_n}$ over parameters $w_1,\dots,w_n$ that maps an element of
$Y$ to the (approximated) probability of $\probvar{B}_x$ being evaluated to $y$.
The parameters are tuned using numerical optimization so that the probability of
$D$ being generated is maximized.
%
%
Then, the function $f$ is synthesized so that $f(x) = \argmax_{y \in Y}
F_{x,w_1,\dots,w_n}(y)$.

The performance of $f$ synthesized as above depends on several factors,
including: (1) whether the set $\{F_{x,w_1,\dots,w_n} \mid \mbox{$w_1,\dots,w_n$
are possible parameters}\}$ contains a function that is sufficiently ``close''
to the true distribution $d_x$ and (2) $x \mapsto F_{x,w_1,\dots,w_n}$ is not
overfitted to $D$ and performs well for unobserved data.
Deep learning, as seen in \sect{back-deep}, is a promising methodology
to address these issues.







%

\section{Automated proof synthesis with statistical modeling}
\label{sec:proofConstruction}

This section starts with making a statistical model for the APS problem by
rephrasing the concepts introduced in \sect{statisticalModeling} using the terms
of our setting.
Based on this model, we define a proof-search procedure that takes the
likelihood of a proof term into account.
We show that we can decompose the probability distribution $p( \boldsymbol{M_P} | \boldsymbol{P} )$ for APS to
the multiplication of easier-to-approximate fine-grained probability
distributions in terms of term constructors that occur in proof terms.
%
We show the derivation of these fine-grained distributions and then
proceed to the definition of the procedure of proof synthesis.

\subsection{Automated proof synthesis, statistically}
\label{sec:back-ATP}

In our setting, the training dataset $D$ is a set of
proposition--proof pairs
$\{(\ottnt{P'_{{\mathrm{1}}}},\ottnt{M'_{{\mathrm{1}}}}),\dots,(\ottnt{P'_{\ottmv{n}}},\ottnt{M'_{\ottmv{n}}})\}$, where
$\ottnt{M'_{\ottmv{i}}}$ is a complete proof of $\ottnt{P'_{\ottmv{i}}}$ for each $i$.
We are to synthesize a function $f$ that takes a proposition $\ottnt{P}$
and returns its proof $\ottnt{M}$ for as many propositions $\ottnt{P}$ as
possible, that is, our $f$ has to satisfy $ \emptyset   \vdash   f (  \ottnt{P}  )  \,  \mathrel{:}  \, \ottnt{P}$ for
many propositions $\ottnt{P}$.

%
%

To statistically model the APS problem, we designate several random
variables.  $ \boldsymbol{P} $ is a random variable that evaluates to a
type.\footnote{
  We could actually build a statistical model that does not treat
  $ \boldsymbol{P} $ as a random variable as long as we focus on the
  contents of this paper.  However, in order to easily extend our
  framework to one that consider nontrivial probabilistic
  distributions over propositions in future, we model $ \boldsymbol{P} $ as
  a random variable.}  $ \boldsymbol{M_P} $ is a family of random
variables indexed by type $\ottnt{P}$.
%
Then, for a predicate $\probvar{\varphi}( \boldsymbol{P} )$ on
$ \boldsymbol{P} $, the probability distribution
$p( \boldsymbol{M_P}  \mid \probvar{\varphi})$ is a
distribution on the set of proof terms conditioned by the predicate
$\probvar{\varphi}( \boldsymbol{P} )$.
As we outlined in Section~\ref{sec:statisticalMachineLearning}, if we
have a good approximation of the conditional probability distribution
$\mathit{p}  \ottsym{(}  \boldsymbol{M_P}  \mid   \boldsymbol{P}  =  \ottnt{P}   \ottsym{)}$, then we obtain a
highly probable proof term of $\ottnt{P}$ by computing
$\argmax_{\ottnt{M}} \mathit{p}  \ottsym{(}   \boldsymbol{M_P}  =  \ottnt{M}   \mid   \boldsymbol{P}  =  \ottnt{P}   \ottsym{)}$.
Therefore, a function that maps $\ottnt{P}$ to
$\argmax_{\ottnt{M}} \mathit{p}  \ottsym{(}   \boldsymbol{M_P}  =  \ottnt{M}   \mid   \boldsymbol{P}  =  \ottnt{P}   \ottsym{)}$ is
expected to be a good proof synthesizer under this model.
%
%
%
%
%
%


\subsection{Derivation of fine-grained distributions}
\label{sec:finegrained}

We could directly learn $\mathit{p}  \ottsym{(}  \boldsymbol{M_P}  \mid   \boldsymbol{P}  =  \ottnt{P}   \ottsym{)}$ using a certain machine
leaning technique along with the training dataset $D$; this is the
strategy taken by \citet{DBLP:journals/corr/SekiyamaIS17}.
However, we found that such monolithic approximation of the
probabilistic distribution often leads to a bad approximation; indeed,
the accuracy of the automated proof synthesizer by Sekiyama et al.\
was around $50\%$ at best.
In this paper, we instead convert $\mathit{p}  \ottsym{(}  \boldsymbol{M_P}  \mid   \boldsymbol{P}  =  \ottnt{P}   \ottsym{)}$ using the laws of
probabilities so that the learning task is reduced to a set of
fine-grained ones.
Under this strategy, one can compute an approximation of the
probability distribution by combining several easier-to-approximate
distributions.
We discover that the combination leads to a better proof synthesizer
than learning $\ottnt{M}$ monolithically.
%





In order to derive the fine-grained distributions, we first introduce
several notions that enable us to specify occurrences of term
constructors in proofs.

\begin{definition}[One-depth contexts]
  The set of \emph{one-depth contexts} is defined by the following BNF:
  \[
    \begin{array}{rcl}
      \ottnt{C} \,\mathop{\invin}\, {\CTXS} &::=& \lambda  \mathit{x}  \ottsym{.}  \left[ \, \right] \mid \left[ \, \right] \, \left[ \, \right] \mid
                                              \ottsym{(}  \left[ \, \right]  \ottsym{,}  \left[ \, \right]  \ottsym{)} \mid \mathsf{case} \, \left[ \, \right] \, \mathsf{of} \, \ottsym{(}  \mathit{x}  \ottsym{,}  \mathit{y}  \ottsym{)}  \rightarrow  \left[ \, \right] \mid\\
                                        && \mathsf{Left} \, \left[ \, \right] \mid \mathsf{Right} \, \left[ \, \right] \mid \mathsf{case} \, \left[ \, \right] \, \mathsf{of} \, \ottsym{\{} \, \mathsf{Left} \, \mathit{x}  \rightarrow  \left[ \, \right]  \ottsym{;} \, \mathsf{Right} \, \mathit{y}  \rightarrow  \left[ \, \right]  \ottsym{\}}.
    \end{array}
  \]
  We assume that each hole in a one-depth context is equipped with a
  unique identifier.  We write $ \ottnt{C}  \overline{ [  \ottnt{M_{\ottmv{i}}}  ]_{ \ottmv{i} } } $ for the term obtained by
  filling holes $ \left[ \, \right] _{  0  } , ...,  \left[ \, \right] _{ \ottmv{n} } $ in $\ottnt{C}$ with terms
  $\ottnt{M_{{\mathrm{0}}}}, ..., \ottnt{M_{\ottmv{n}}}$, respectively.
\end{definition}

\begin{definition}[Paths]
  A \emph{path} $\rho$ is a finite sequence of pairs $\ottsym{(}  \ottnt{C}  \ottsym{,}  \ottmv{i}  \ottsym{)}$
  where $i$ is a natural number that identifies a hole in $\ottnt{C}$.
  We write $ \langle  \rho ,  \ottsym{(}  \ottnt{C}  \ottsym{,}  \ottmv{i}  \ottsym{)}  \rangle $ for the path obtained by postpending
  $\ottsym{(}  \ottnt{C}  \ottsym{,}  \ottmv{i}  \ottsym{)}$ to path $\rho$.
\end{definition}

%

A one-depth context represents a term constructor other than
variables.  Using one-depth contexts,
$\rho = \langle \ottsym{(}  \ottnt{C_{{\mathrm{0}}}}  \ottsym{,}   \ottmv{i} _{  0  }   \ottsym{)},\ottsym{(}  \ottnt{C_{{\mathrm{1}}}}  \ottsym{,}   \ottmv{i} _{  1  }   \ottsym{)},\dots,\ottsym{(}  \ottnt{C_{\ottmv{n}}}  \ottsym{,}   \ottmv{i} _{ \ottmv{n} }   \ottsym{)}
\rangle$ specifies a path in a term, whose top-level constructor is
identical to $\ottnt{C_{{\mathrm{0}}}}$, from its root node in the following way:
$\ottnt{C_{{\mathrm{0}}}}$; the hole in $\ottnt{C_{{\mathrm{0}}}}$ with the identifier $ \ottmv{i} _{  0  } $;
$\ottnt{C_{{\mathrm{1}}}}$; the hole in $\ottnt{C_{{\mathrm{1}}}}$ with the identifier $ \ottmv{i} _{  1  } $; and so
on.
%
%
For example, let $\ottnt{M}$ be a term $\lambda  \mathit{x}  \ottsym{.}  \mathsf{case} \, \mathit{x} \, \mathsf{of} \, \ottsym{(}  \mathit{y}  \ottsym{,}  \mathit{z}  \ottsym{)}  \rightarrow  \ottsym{(}  \mathit{z}  \ottsym{,}  \mathit{y}  \ottsym{)}$.
Then, a path from the root of $\ottnt{M}$ to the reference to variable
$\mathit{y}$ is represented by the path
\[
 \langle \ottsym{(}  \lambda  \mathit{x}  \ottsym{.}   \left[ \, \right] _{  0  }   \ottsym{,}   0   \ottsym{)}, \ \ottsym{(}  \mathsf{case} \,  \left[ \, \right] _{  0  }  \, \mathsf{of} \, \ottsym{(}  \mathit{y}  \ottsym{,}  \mathit{z}  \ottsym{)}  \rightarrow   \left[ \, \right] _{  1  }   \ottsym{,}   1   \ottsym{)}, \ \ottsym{(}  \ottsym{(}   \left[ \, \right] _{  0  }   \ottsym{,}   \left[ \, \right] _{  1  }   \ottsym{)}  \ottsym{,}   1   \ottsym{)} \rangle.
\]

We show that the probability $\mathit{p}  \ottsym{(}   \boldsymbol{M_P}  =  \ottnt{M}   \mid   \boldsymbol{P}  =  \ottnt{P}   \ottsym{)}$ is equal to
$\phi(M,  \langle \rangle )$ \emph{if every subterm $\ottnt{M'}$ of $\ottnt{M}$ is
  annotated with its type (which we write $\TYPEOF(\ottnt{M'})$)}, where
$ \phi $ is defined by induction on the structure of $M$:
\begin{equation}
  \label{eq:phiprime}
 \begin{array}{rcl}
   \phi \, \ottsym{(}  \mathit{x}  \ottsym{,}  \rho  \ottsym{)} &=& p(\probvar{x}=x \mid \probvar{P} = P,  \boldsymbol{Q} =\TYPEOF(x), \probvar{\rho} = \rho) \\
   \phi \, \ottsym{(}   \ottnt{C}  \overline{ [  \ottnt{M_{\ottmv{i}}}  ]_{ \ottmv{i} } }   \ottsym{,}  \rho  \ottsym{)}  &=&
                                           \begin{array}[t]{@{}l}
                                             p(\probvar{C}=C \mid \probvar{P}=P,  \boldsymbol{Q} =\TYPEOF( \ottnt{C}  \overline{ [  \ottnt{M_{\ottmv{i}}}  ]_{ \ottmv{i} } } ), \probvar{\rho}=\rho) \mathop{\times}\\
                                             \prod_{i}
                                             \phi \, \ottsym{(}  \ottnt{M_{\ottmv{i}}}  \ottsym{,}    \langle  \rho ,  \ottsym{(}  \ottnt{C}  \ottsym{,}  \ottmv{i}  \ottsym{)}  \rangle    \ottsym{)}.
                                             \end{array}
 \end{array}
\end{equation}

The function $\phi \, \ottsym{(}  \ottnt{M}  \ottsym{,}  \rho  \ottsym{)}$ computes
$p(\probvar{M}=M | \probvar{P}=P,  \boldsymbol{Q} =\TYPEOF(M),
\probvar{\rho}=\rho)$, the probability of $M$ being a subterm of $\TYPEOF(M)$ at the position specified by $\rho$ within a proof of $\ottnt{P}$, by
induction on the structure of $M$ using two auxiliary probabilities:
$p(\probvar{x}=x \mid \probvar{P}=P,  \boldsymbol{Q}  =  \ottnt{Q} , \probvar{\rho} =
\rho)$ and
$p(\probvar{C}=C \mid \probvar{P}=\ottnt{P},  \boldsymbol{Q}  =  \ottnt{Q} ,
\probvar{\rho}=\rho)$.
In the definition, we use the following random
variables: $\probvar{x}$ evaluates to a term variable; $\probvar{C}$
evaluates to a one-depth context;
$ \boldsymbol{Q} $ evaluates to the type to be proved by $M$;
and $ \boldsymbol{\rho} $ evaluates to a path that specifies the position where $\mathit{x}$ or $\ottnt{C}$ is placed.  Note that $ \boldsymbol{P} $ evaluates to the
type that is \emph{supposed to be proved by the root node}, not by
$\ottnt{M}$.
%
%
The conditional-probability expression
$p(\probvar{x}=x \mid \probvar{P}=P,  \boldsymbol{Q}  =  \ottnt{Q} , \probvar{\rho} =
\rho)$ quantifies the probability of $x$ being a proof of $Q$
under the condition that it appears at the position specified by
$\rho$; and
$p(\probvar{C}=C \mid \probvar{P}=\ottnt{P},  \boldsymbol{Q}  =  \ottnt{Q} ,
\probvar{\rho}=\rho)$ is the probability of $C$ being the
top-level constructor of a proof term of $\ottnt{Q}$ if it appears at the
position specified by $\rho$.
%
We will explain how to model type annotations $\TYPEOF(\ottnt{M})$ later.


If $M$ is a variable $x$, then $\phi$ uses the value of the former
probability as the answer.  If $M$ is not a variable, then it can be
written in the form of $ \ottnt{C}  \overline{ [  \ottnt{M_{\ottmv{i}}}  ]_{ \ottmv{i} } } $ using some one-depth context $C$
and terms $M_1,\dots,M_n$.  The definition argues that this
probability is the multiplication of
(1) the likelihood of $C$ conditioned by $ \boldsymbol{P}  =  \ottnt{P} $, $\probvar{Q}=\TYPEOF(M)$, and $\probvar{\rho}=\rho$; and
(2) the probabilities of
$\phi(M_i,  \langle  \ottsym{(}  \ottnt{C}  \ottsym{,}  \ottmv{i}  \ottsym{)} ,  \rho  \rangle )$.  Notice that the information of $M_j$
does \emph{not} appear in the probability calculation of $M_k$ if
$j \ne k$; this greatly simplifies the definition of $\phi$.

We informally show that $\phi(M, \rho)$ is indeed equal to
$p(\probvar{M}=M \mid \probvar{P}=P,  \boldsymbol{Q} =\TYPEOF(M),
\probvar{\rho}=\rho)$ by induction on the structure of
$M$.\footnote{In order to formally prove this fact, we need to define
  the random variables and the probability distributions in this paper
  in more formal style, which we decide to defer to future work.}  The
case of $M=x$ is easy.  Consider the case of $\ottnt{M} \, \ottsym{=} \,  \ottnt{C}  \overline{ [  \ottnt{M_{\ottmv{i}}}  ]_{ \ottmv{i} } } $.  We
start from
$p(\probvar{M}= \ottnt{C}  \overline{ [  \ottnt{M_{\ottmv{i}}}  ]_{ \ottmv{i} } }  \mid \probvar{P}=P,  \boldsymbol{Q}  =  \textsf{typeof} \, \ottsym{(}  \ottnt{M}  \ottsym{)} ,
\probvar{\rho}=\rho)$.  This probability is equal to the following
probability:
\[
  p\left(\probvar{C}=C, \probvar{M_1}=\ottnt{M_{{\mathrm{1}}}}, \dots, \probvar{M_n}=\ottnt{M_{\ottmv{n}}} \ \middle|
  \begin{array}{l@{}}
    \probvar{P}=P,  \boldsymbol{Q}  =  \textsf{typeof} \, \ottsym{(}  \ottnt{M}  \ottsym{)} , \probvar{\rho}=\rho
  \end{array}
    \right).
  \]
  Here,
  each $\probvar{M}_i$ is the random variable that
  evaluates to the term to be filled in the $i$-th hole in $C$.
  By using the chain rule of conditional
  probabilities~\cite{Koller:2009:PGM:1795555}
  this probability is equal to the following.
  \begin{equation}
    \label{eq:jointProb}
    \begin{array}{l}
      p\left(\probvar{C}=C \mid
        \probvar{P}=P,  \boldsymbol{Q} =\TYPEOF(M), \probvar{\rho}=\rho
      \right) \times\\
      p\left(\probvar{M_1}=M_1, \dots, \probvar{M_n}=M_n \mid
        \probvar{C}=C,
      \probvar{P}=P,  \boldsymbol{Q} =\TYPEOF(M), \probvar{\rho}=\rho 
      \right).
    \end{array}
  \end{equation}

  We decompose the second expression in Equation~\ref{eq:jointProb}.
  Let us first refine the condition of this expression with
  $\probvar{Q}_i$, a family of the random variables that evaluate to
  the type of the $i$-th hole in $C$, and $\probvar{\rho}_i$, a family
  of random variables that evaluate to the paths of the $i$-th hole in
  $C$.  For short, let us write
  $\probvar{\overrightarrow{Q}}_i=\TYPEOF(\overrightarrow{M_i})$ for
  $\probvar{Q}_1=\TYPEOF(M_1), ..., \probvar{Q}_n=\TYPEOF(M_n)$ and
  $\probvar{\overrightarrow{\rho_i}}=\overrightarrow{\rho_i}$ for
  $\probvar{\rho}_1=\langle\rho,(C,1)\rangle, ...,
  \probvar{\rho}_n=\langle\rho,(C,n)\rangle$.  Then, with these random
  variables, the second expression in Equation~\ref{eq:jointProb} is
  equal to
  \[
    p\left(\probvar{M_1}=M_1, \dots, \probvar{M_n}=M_n \middle|
      \probvar{C}=C,
      \probvar{P}=P, \probvar{Q}=\TYPEOF(M), \probvar{\rho}=\rho, 
      \probvar{\overrightarrow{Q}}_i=\TYPEOF(\overrightarrow{M_i}),
      \probvar{\overrightarrow{\rho}}_i=\overrightarrow{\rho_i}
      \right)
    \]
    because (1) the condition
    $\probvar{\overrightarrow{Q}}_i=\TYPEOF(\overrightarrow{M_i})$
    does not affect the distributions of
    $\probvar{M_1}, \dots, \probvar{M_n}$ since the type of every
    subexpression is known by assumption and (2) the condition
    $\probvar{\overrightarrow{\rho}}_i=\overrightarrow{\rho_i}$ does
    not affect the distributions either since the values of
    $\probvar{\overrightarrow{\rho}}_i$ are uniquely determined by
    $C$ and $\rho$.  Then, an important observation here is that
    \emph{under the condition that the values of
      $\probvar{\overrightarrow{Q}_i}$ and $\probvar{\overrightarrow{\rho}_i}$
      are known, each of the random variables
      $\probvar{M}_1, \dots, \probvar{M}_n$ is independent of the
      rest}; in other words, each of the random variables
    $\probvar{M}_1, \dots, \probvar{M}_n$ is \emph{conditionally
      independent} of the rest under
      $\probvar{\overrightarrow{Q}_i}$ and $\probvar{\overrightarrow{\rho}_i}$.
      Hence, the above probability is equal to
  \[
    \prod_{1 \le j \le n}
    p\left(\probvar{M}_j=M_j \middle|
            \probvar{C}=C,
      \probvar{P}=P, \probvar{Q}=\TYPEOF(M), \probvar{\rho}=\rho, 
      \probvar{\overrightarrow{Q}}_i=\TYPEOF(\overrightarrow{M_i}),
      \probvar{\overrightarrow{\rho}}_i=\overrightarrow{\rho_i}
    \right).
  \]
  In the condition part of this expression, noting that (1) the values
  of $ \boldsymbol{C} $, $\probvar{Q}$, and $ \boldsymbol{\rho} $ are irrelevant since we have the
  values of $\probvar{\overrightarrow{Q}}_i$ and
  $\probvar{\overrightarrow{\rho}}_i$; and (2) the distribution of
  $\probvar{M}_j$ depends only on the values of $\probvar{\rho}_j$ and
  $\probvar{Q}_j$, the above probability is equal to
    \[
    \prod_{1 \le j \le n}
    p\left(\probvar{M}_j=M_j \middle|
      \probvar{P}=P,
      \probvar{Q}_j=\TYPEOF(M_j),
      \probvar{\rho}_j=\rho_j
    \right).
  \]
  By the induction hypothesis, this is
  equal to $\prod_{1 \le j \le n}\phi(M_j, \langle\rho,(C,j)\rangle)$;
  substituting this to Equation~\ref{eq:jointProb}, we have
  Equation~\ref{eq:phiprime}.
  
  We back up the aforementioned observation that is a key to decompose
the second expression by an example.  Let $C$ be $\ottsym{(}   \left[ \, \right] _{  0  }   \ottsym{,}   \left[ \, \right] _{  1  }   \ottsym{)}$
under a path that binds $f$ to $\mathit{a} \,  \mathord{\rightarrow}  \, \mathit{b}$, $g$ to $\mathit{a} \,  \mathord{\times}  \, \mathit{b} \,  \mathord{\rightarrow}  \, \mathit{a}$, $x$ to
$\mathit{a}$, and $y$ to $\mathit{b}$.  Suppose we know that $ \left[ \, \right] _{  0  } $ should be
filled with a term of type $\mathit{a} \,  \mathord{\rightarrow}  \, \mathit{b}$ and $ \left[ \, \right] _{  1  } $ with a term of
type $\mathit{a}$.
Obviously, in our context of proof
  synthesis, the fact that $f$ is filled in $ \left[ \, \right] _{  0  } $ does not add any
  restriction on the set of possible terms in $ \left[ \, \right] _{  1  } $, since
  \emph{any} term of the type of each hole works as a proof of each type.
  This observation can be generalized to an arbitrary case in our type
  system.\footnote{We also expect this observation to be generalized
    to various type systems.  This observation essentially comes from
    the fact that the interfacing by a type separates certain
    dependency between a context and a term, which is often true for
    many type systems.}

\subsection{Proof synthesis procedure}
\label{sec:proof-synthesis}

\begin{algorithm}[t]
 \caption{Proof synthesis}
 \label{alg:proof}
 \begin{algorithmic}[1]
  \Procedure{ProofSynthesize}{$\ottnt{P}$}
  \State Initialize priority queue {\queue} that contains partial proofs
         constructed so far.
  \State Push $ \left[ \, \right] $ to {\queue} with priority $1.0$.
  \While{{\queue} is not empty} \label{alg:proof:main-loop}
    \State Pop $\ottnt{M}$ with the highest priority {\prob} from {\queue}.
    \State Let $\rho = \argmax_{\rho \in \textsf{hole} \, \ottsym{(}  \ottnt{M}  \ottsym{)}} \max_{\mathit{r}}  \mathit{p} ^*   \ottsym{(}   \boldsymbol{r}  =  \mathit{r}   \mid   \boldsymbol{\rho}  =  \rho   \ottsym{)}$.
    \For{each $\mathit{C_x} \in {\CTXS} \cup \textsf{BV} \, \ottsym{(}  \ottnt{M}  \ottsym{,}  \rho  \ottsym{)}$ such that $ \emptyset   \vdash   \ottnt{M}  [  \mathit{C_x}  ]_{ \rho }  \,  \mathrel{:}  \, \ottnt{P}$} \label{alg:proof:typechecking}
      \If{$\textsf{hole} \, \ottsym{(}   \ottnt{M}  [  \mathit{C_x}  ]_{ \rho }   \ottsym{)} = \emptyset$}
        \State \Return $ \ottnt{M}  [  \mathit{C_x}  ]_{ \rho } $ \label{alg:proof:return}
      \Else
        \State Let $\ottnt{Q}$ be a proof obligation to be discharged at $ \left[ \, \right] _{ \rho } $.
        \State Push $ \ottnt{M}  [  \mathit{C_x}  ]_{ \rho } $ to {\queue} with priority $ \mathit{p} ^*   \ottsym{(}   \boldsymbol{r}  =   \mathit{r} _{ \mathit{C_x} }    \mid   \boldsymbol{P}  =  \ottnt{P}   \ottsym{,}   \boldsymbol{\rho}  =  \rho   \ottsym{,}   \boldsymbol{Q}  =  \ottnt{Q}   \ottsym{)}\,{\prob}$ \label{alg:proof:add}
      \EndIf
    \EndFor
  \EndWhile \label{alg:proof:main-loop-end}
  \EndProcedure
 \end{algorithmic}
\end{algorithm}

Based on the discussion in Section~\ref{sec:finegrained}, we design a
proof-synthesis procedure.
\proc{proof} shows the definition of our procedure
\textsc{ProofSynthesize}, which takes proposition $\ottnt{P}$ to be
proved.
This procedure maintains a priority queue {\queue} of partially constructed terms.
%
The priority associated with $\ottnt{M}$ by {\queue} denotes the
likelihood of $\ottnt{M}$ forming a proof of $\ottnt{P}$.
In each iteration of
Lines~\ref{alg:proof:main-loop}--\ref{alg:proof:main-loop-end},
\textsc{ProofSynthesize} picks a term $\ottnt{M}$ with the highest
likelihood and fills a hole in $\ottnt{M}$ with a one-depth context.
It returns a proof if it encounters a correct proof of $\ottnt{P}$.
We write $ \mathit{p} ^*   \ottsym{(}   \boldsymbol{\varphi} _{  1  }   \mid   \boldsymbol{\varphi} _{  2  }   \ottsym{)}$ for an approximation of
$\mathit{p}  \ottsym{(}   \boldsymbol{\varphi} _{  1  }   \mid   \boldsymbol{\varphi} _{  2  }   \ottsym{)}$.



Before going into the detail, we remark a gap between the procedure
\textsc{ProofSynthesize} and the statistical model in
\sect{finegrained}.  In that statistical model, we defined the
likelihood of a variable
$p(\probvar{x} \mid \probvar{P}=P,  \boldsymbol{Q}  =  \ottnt{Q} , \probvar{\rho} =
\rho)$ and that of a one-depth context
$p(\probvar{C} \mid \probvar{P}=\ottnt{P},  \boldsymbol{Q}  =  \ottnt{Q} ,
\probvar{\rho}=\rho)$ as separate probability distributions.
Although this separation admits the inductive definition of the
function $\phi$, it is not necessarily plausible from the viewpoint of
proof synthesis since, in filling a hole, we do not know whether it
should be filled with a variable or with a one-depth context.

In order to solve this problem, we assume that we have an
approximation of the likelihood of an \emph{proof inference rule} that should be applied to a hole.
Concretely, we assume that we can approximate the probability
distribution $\mathit{p}  \ottsym{(}  \boldsymbol{r}  \mid  \boldsymbol{P}  \ottsym{,}  \boldsymbol{\rho}  \ottsym{,}  \boldsymbol{Q}  \ottsym{)}$, where $ \boldsymbol{r} $ is a random
variable that evaluates to the name of an proof inference rule in
Figure~\ref{fig:infrule}.  This assumption requires that we
estimate the likelihood of $\ottdrulename{Var}$ being applied for a
hole, which can be done in the same way as estimation of those of other inference rules.


%
Let us explain the inside of the procedure in more detail.
A proof is synthesized by the while loop, where the procedure fills
the hole $ \left[ \, \right] _{ \rho } $ pointed by $\rho$ in the partial proof
$\ottnt{M}$ that has the highest likelihood $\prob$
(Lines~\ref{alg:proof:main-loop}--\ref{alg:proof:main-loop-end}).
We write $\textsf{hole} \, \ottsym{(}  \ottnt{M}  \ottsym{)}$ for the set of paths to holes in $\ottnt{M}$.
We select path $\rho$ such that the inference rule applied at the
position pointed by the path has the highest probability.
After finding the hole to be filled, we replace it with $\mathit{C_x}$,
which denotes one-depth contexts or variables.
$\textsf{BV} \, \ottsym{(}  \ottnt{M}  \ottsym{,}  \rho  \ottsym{)}$ is the set of bound variables that can be referred
to at $ \left[ \, \right] _{ \rho } $ and $ \ottnt{M}  [  \mathit{C_x}  ]_{ \rho } $ is the term obtained by filling
$ \left[ \, \right] _{ \rho } $ in $\ottnt{M}$ with $\mathit{C_x}$.
Note that {\CTXS} are the set of all one-depth contexts.
If $ \ottnt{M}  [  \mathit{C_x}  ]_{ \rho } $ is a proof of $\ottnt{P}$, which can be checked using
an off-the-shelf type checker, then the procedure returns it as the
synthesis result (Line~\ref{alg:proof:return}).
Otherwise, $ \ottnt{M}  [  \mathit{C_x}  ]_{ \rho } $ is added to {\queue} with priority
$ \mathit{p} ^*   \ottsym{(}   \boldsymbol{r}  =   \mathit{r} _{ \mathit{C_x} }    \mid   \boldsymbol{P}  =  \ottnt{P}   \ottsym{,}   \boldsymbol{\rho}  =  \rho   \ottsym{,}   \boldsymbol{Q}  =  \ottnt{Q}   \ottsym{)}\,{\prob}$, which is the
likelihood of $ \ottnt{M}  [  \mathit{C_x}  ]_{ \rho } $ forming a proof
(Line~\ref{alg:proof:add}).
$ \mathit{r} _{ \mathit{C_x} } $ is the proof inference rule corresponding to $\mathit{C_x}$.
$\ottnt{Q}$ is a proof obligation at $ \left[ \, \right] _{ \rho } $; how to find it is
discussed in \sect{exp-proof}.

We make a few remarks about the procedure:
\begin{itemize}
\item In the current implementation, we have not implemented the
  approximator of $\mathit{p}  \ottsym{(}  \boldsymbol{x}  \mid  \boldsymbol{P}  \ottsym{,}  \boldsymbol{\rho}  \ottsym{,}  \boldsymbol{Q}  \ottsym{)}$; instead, if the procedure
  decides to fill a hole with a variable, we assume that
  $\mathit{p}  \ottsym{(}  \boldsymbol{x}  \mid  \boldsymbol{P}  \ottsym{,}  \boldsymbol{\rho}  \ottsym{,}  \boldsymbol{Q}  \ottsym{)}$ is the uniformly distribution on the set of
  variables that are available at this scope.  Although this may look
  like a naive strategy, our implementation still works quite well for
  many propositions; see \sect{exp-proof}.  The problem of estimating
  the likelihood of a variable is similar to the \emph{premise selection
    problem}, for which various work has been
  done~\cite{Irving_NIPS_2016,Kaliszyk_arxiv_2017,Wang_2017_NIPS,DBLP:conf/lpar/LoosISK17}.
  Combining our synthesizer with such a technique is an interesting
  future direction.

  
\item In the current implementation, we assume that the type checking
  conducted in Line~\ref{alg:proof:typechecking} infers the type of
  each subexpression of $ \ottnt{M}  [  \mathit{C_x}  ]_{ \rho } $ and annotates these types to
  them; this is indeed how we handle the $\textsf{typeof} \, \ottsym{(}  \ottnt{M}  \ottsym{)}$ in
  \sect{finegrained}.  This is a reasonable assumption as far as
  we are concerned with the propositional logic.  For more expressive
  logics, we may need some auxiliary methods to guess the type of each
  expression.
\item The procedure \textsc{ProofSynthesize} is not an algorithm.  If
  it is fed with an unsatisfiable proposition, then it does not
  terminate.  Even if it is fed with a valid proposition, it may not
  be able to discover a proof of the proposition depending on the
  performance of the estimator of $ \mathit{p} ^* $.
  %
\end{itemize}

\section{Neural proposition-to-proof model}
\label{sec:model-deep}

In order to implement \textsc{ProofSynthesize}, we are to approximate the
probability distribution $\mathit{p}  \ottsym{(}  \boldsymbol{r}  \mid  \boldsymbol{P}  \ottsym{,}  \boldsymbol{\rho}  \ottsym{,}  \boldsymbol{Q}  \ottsym{)}$ that produces the likelihood of a
proof inference rule being applied at a given position in a proof.
To this end, we design a new DNN model, which we call a
\emph{proposition-to-proof model}, tailored to the classification task of
inference rules.
We start with a brief review of deep learning for making this paper
self-contained; see, e.g., \citet{Goodfellow-et-al-2016} for the details.
Then we describe a basic architecture of the proposition-to-proof model.
%

\subsection{Deep learning, briefly}
\label{sec:back-deep}



Deep learning is a generic term for machine learning methods based on deep
neural networks.
Usually, as other machine learning technologies, the first common step of deep
learning is to build a vector space of \emph{features} and a mapping from a
datum to the space of the features.
%
%
An element of the vector space of the features is called a
\emph{feature vector}.
It is a multidimensional vector each dimension of which is a numerical
value that represents certain information of the datum.
By embedding data to a vector space and working on this space, we can
apply various numerical optimization techniques of machine learning
such as support vector machine and random forest.
%
The design of the feature representation is known to have a great
influence on the performance of a machine learning method.
However, developing feature representations, known as feature-engineering,
requires deep expertise in the application domains; it is in general difficult
to generalize a design of a feature representation to other tasks.

Deep learning allows us to avoid hard feature-engineering.
%
In contrast with other methods, deep learning can learn the feature
representation from data without manual engineering, which makes it
possible to find a good feature representation with less effort.
%
Deep learning is also very expressive in the sense that it can approximate any
continuous function $f$ with a given degree of accuracy by appropriately tuning
the set of parameters~\cite{Hornik_1989_NN}.
Furthermore, it is known that deep learning tends to \emph{generalize}
to unseen data without overfitting to a given training dataset if the
set is sufficiently large.\footnote{The reason why a deep learning
  model tends not to overfit is still not fully understood;
  see~\citet{DBLP:conf/nips/NeyshaburBMS17} for recent remarkable development on
  this issue.}


Models in deep learning are represented by (artificial) neural networks, which
consist of an \emph{input layer} that converts data to a feature
vector; an \emph{output layer} that converts a feature vector to a
human-readable format; and one or more \emph{hidden layers} that learn
feature representations.
Neural networks with multiple hidden layers are especially called
\emph{deep neural networks (DNNs)}.
The basic building block of hidden layers is a \emph{perceptron},
which is a function over multidimensional vectors with learnable
parameters.
Given an $n$-dimensional vector $ \mathit{v} _{ \mathit{x} }  = [x_1, ..., x_n]$ in the
vector space $\RealVec{n}$ on reals, a perceptron produces an
$m$-dimensional vector $ \mathit{v} _{ \mathit{y} }  = [y_1, ..., y_m] \in \RealVec{m}$
such that:
\[
 y_j = \sum_{i = 1}^{n} W_{j,i}\, x_i + b_j
\]
where $W_{j,i} \in \Real$, a matrix called a \emph{weight}, is a
learnable coefficient parameter; and $b_j \in \Real$, a vector called
a \emph{bias}, is a learnable parameter independent of the input.
We simply write the behavior of a perceptron as the following linear
operation:
\[  \mathit{v} _{ \mathit{y} }  = W  \mathit{v} _{ \mathit{x} }  + b
\] where $W$ is a real matrix in $\RealMat{m}{n}$, $b$ is a real
vector in $\RealVec{m}$, $W  \mathit{v} _{ \mathit{x} } $ is the matrix product of $W$
and the transpose of $ \mathit{v} _{ \mathit{x} } $, and $+$ is the element-wise
addition.
Since a perceptron is a linear function, any multilayer perceptron is
also represented by a single perceptron since the composition of
several linear maps is also linear.
To give DNN models the ability to approximate any \emph{nonlinear}
function, each hidden layer postpends the application of a nonlinear
function $f$, called \emph{activation}, and produces the result
$ \mathit{v} _{ \mathit{z} } $ of an element-wise application of $ \mathit{v} _{ \mathit{y} } $ to $f$:
\[
   \mathit{v} _{ \mathit{z} }  = f( \mathit{v} _{ \mathit{y} } ).
\]
A hidden layer of this type is called a \emph{fully connected layer}.
We write $W$ and $b$ in it as {\Wparam{fc}{}} and {\bparam{fc}{}} for
clarification.
Hidden layers are supposed to extract abstract features of data and a
use of multiple layers makes DNNs powerful.
However, training a model with an excessive number of layers requires
expensive learning cost.
This is one of the reasons why many variants of DNNs have been studied
for effective learning; this work can be seen as a new DNN model
tailored to proof synthesis.

As other machine learning methods, deep learning requires
training to tune the learnable parameters.
The parameters are tuned so that a model approximates the distribution
of a given dataset as closely as possible using numerical optimization
techniques such as stochastic gradient descent~\cite{sgd}.
These numerical optimization methods adjust learnable parameters so
that the difference between an expected output and
the actual response from the model is minimized; this difference is
called a \emph{loss value} and a function to calculate loss values is
called a \emph{loss function}.
Splitting a training dataset into multiple small collections called
mini-batches is a popular strategy in the training of a DNN.
In this strategy, the numerical optimization is applied to each
mini-batch to update the learnable parameters.
After the training, one evaluates the performance of the trained model by using
a \emph{validation dataset}, which is different but supposed to origin from the
same distribution as the training dataset.

\subsection{Proposition-to-proof model}
We design a DNN model that takes three arguments, proposition $\ottnt{P}$ to be
proven, path $\rho$ pointing to the hole to be filled, and proof obligation
$\ottnt{Q}$ a term of which should be placed in the hole, and approximates the
probability of a proof inference rule being applied at the position specified
by $\rho$ in a proof of $\ottnt{P}$.
Following the standard manner in deep learning, the model represents features of
the three arguments as real vectors and then approximates the likelihood of each
proof inference rule with them.
We first explain how we learn feature representations of propositions $\ottnt{P}$
and $\ottnt{Q}$.
The features of the path are obtained by extracting those of $\ottnt{P}$ along the
path.
We finally integrate all features into a single feature vector and use it to
estimate a proof inference rule that should be applied.

\subsubsection{Proposition encoder}
\label{sec:model-deep-encoder}
To obtain informative features from proposition $\ottnt{P}$, we consider an abstract
syntax tree (AST) representation of $\ottnt{P}$.
Each node of the AST is equipped with a proposition constructor ($ \rightarrow $,
$ \mathord{\times} $, or $ \mathord{+} $) or a propositional variable.
We first give a simple feature to each node in the AST and then design a new
layer to learn an effective feature representation of the AST.
In what follows, we suppose that each node in an AST is associated with a
feature vector.

One possible way to provide vectors that distinguish nodes of an AST is to use
one-hot vectors, which are used broadly in natural language processing and
represent a word as an $n$-dimensional vector ($n$ is the number of unique words
considered) that only the element corresponding to the word has scalar value $1$
and the others have $0$.
In this work, the information of proposition constructor is embedded
into a vector as in one-hot vectors, while propositional variables
embed their numerical scale values into a fixed element in the vector.
\begin{definition}[Vector representation of proposition node]
 Let $f$ be a bijective function that maps propositional variables to positive
 numbers.
 Then, $ \textsf{Enc} $ gives a vector to node $\mathit{t}$ as follows.
 \[\begin{array}{lll}
  \textsf{Enc} \, \ottsym{(}  \mathit{a}  \ottsym{)}    &=&  [   f (  \mathit{a}  )   \ottsym{,}   0   \ottsym{,}   0   \ottsym{,}   0   ]  \\
  \textsf{Enc} \, \ottsym{(}   \mathord{\rightarrow}   \ottsym{)}   &=&  [   0   \ottsym{,}   1   \ottsym{,}   0   \ottsym{,}   0   ]  \\
  \textsf{Enc} \, \ottsym{(}   \mathord{\times}   \ottsym{)}    &=&  [   0   \ottsym{,}   0   \ottsym{,}   1   \ottsym{,}   0   ]  \\
  \textsf{Enc} \, \ottsym{(}   \mathord{+}   \ottsym{)}    &=&  [   0   \ottsym{,}   0   \ottsym{,}   0   \ottsym{,}   1   ]  \\
   \end{array}\]
\end{definition}
\begin{figure}[t]
 \includegraphics[width=\textwidth]{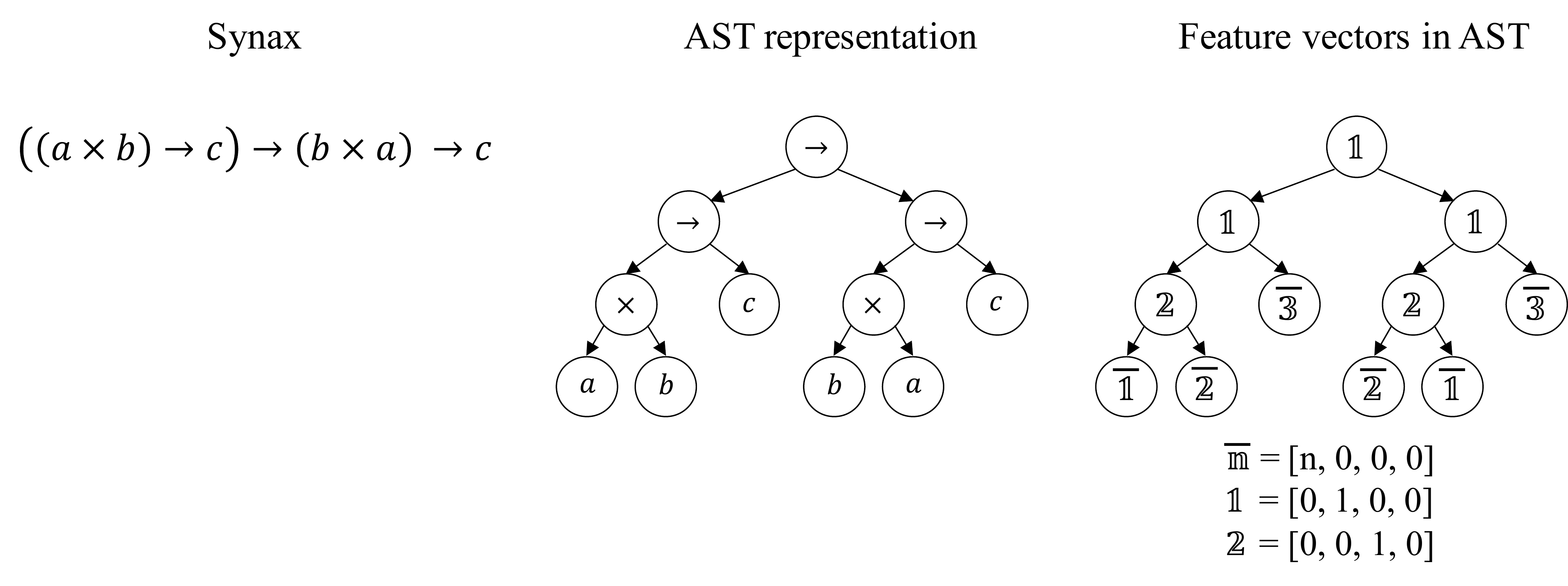}
 \caption{Proposition representations.}
 \label{fig:prop-encode}
\end{figure}
\fig{prop-encode} illustrates feature vectors given by $ \textsf{Enc} $, which are
similar to one-hot vectors in that each dimension of them represents a class of
a node.
We consider that all propositional variables belong to the same class, so
$ \textsf{Enc} $ assigns their numerical values to the same dimension.
On the other hand, different propositional variables should be distinguished;
e.g., if $\mathit{a} \, \neq \, \mathit{b}$, proofs generated for $\mathit{a} \,  \mathord{\rightarrow}  \, \mathit{b} \,  \mathord{\rightarrow}  \, \mathit{a}$ and $\mathit{a} \,  \mathord{\rightarrow}  \, \mathit{b} \,  \mathord{\rightarrow}  \, \mathit{b}$
should be different.
Thus, $ \textsf{Enc} $ assigns different numbers to different propositional variables.

We expect that this encoding of nodes is more informative, especially, for
propositional variables that do not occur in a training dataset---we call such
variables unknown---than one-hot vectors.
How to handle unknown entities is a common issue also in natural language
processing, which addresses the issue by a workaround that maps all unknown
words to a special symbol ``unknown''.
However, this workaround has the problems that (1) the feature vector for the
``unknown'' is not related to propositional variables at the training phase
since the ``unknown'' does not occur in the training dataset and (2) all
unknown propositional variables are mapped to a single symbol ``unknown'' and so
they are not distinguished.
Fortunately, we know as a domain knowledge that ``unknown'' comes from only
propositional variables and we can assign unique positive numbers to all
propositional variables, which should make feature vectors of unknown
propositional variables more informative.
We expect the encoding by $ \textsf{Enc} $ to be helpful for the issue of unknown
propositional variables.

\begin{figure}[t]
 \includegraphics[width=\textwidth]{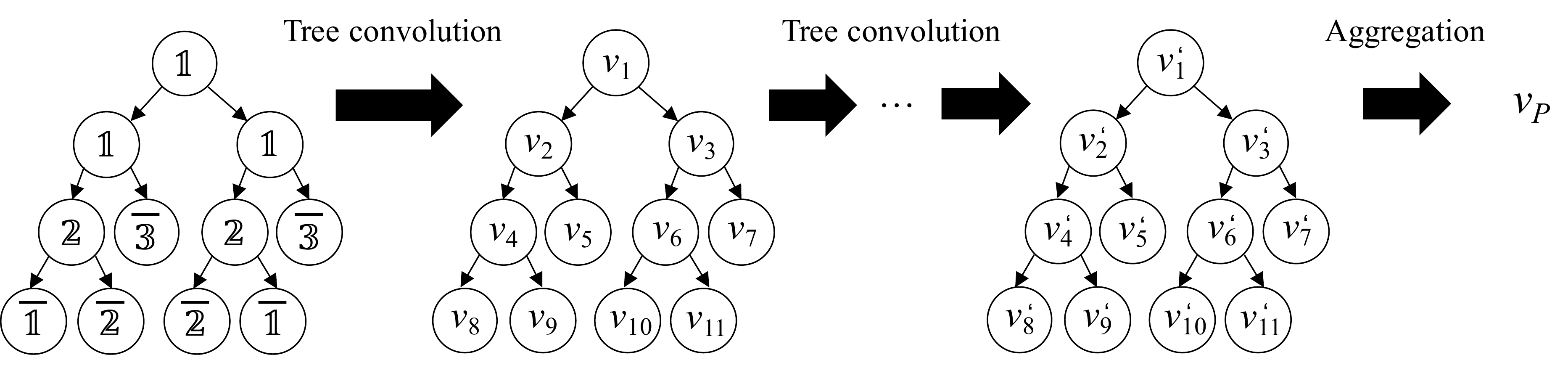}
 \caption{Encoder.}
 \label{fig:encoder}
\end{figure}
After giving a vector to each node by $ \textsf{Enc} $, we obtain features of $\ottnt{P}$
by two steps (\fig{encoder}).
The first step gains a feature representation of each node from nodes around
it by using \emph{AST convolution layers}.
The second step aggregates feature vectors of nodes into a single vector by an
\emph{aggregation layer}.
\paragraph{AST convolution layer}
An AST convolution layer updates a feature vector of each node $\mathit{t}$ in an
AST by using vectors of nodes around $\mathit{t}$.
Suppose that $\textsf{parent} \, \ottsym{(}  \mathit{t}  \ottsym{)}$ is the parent and $\textsf{child} \, \ottsym{(}  \mathit{t}  \ottsym{,}  \ottmv{i}  \ottsym{)}$ is the
$i$-th child of $\mathit{t}$.
Let $ \mathit{v} _{ \mathit{t} } $ be an $n$-dimensional feature vector of $\mathit{t}$.
Then, the AST convolution layer updates all vectors of nodes in a given AST
simultaneously as follows.
Let $ \varsigma _{ \mathit{t} } $ be a class of node $\mathit{t}$, that is, a proposition
constructor ($ \rightarrow $, $ \mathord{\times} $, or $ \mathord{+} $) or a class to denote propositional
variables.
\begin{equation}
  \mathit{v} _{ \mathit{t} }  \leftarrow {\Fact{conv}} \left(
   \sum_{i} \Wparam{conv}{ \varsigma _{ \mathit{t} } ,i} \,  \mathit{v} _{ \textsf{child} \, \ottsym{(}  \mathit{t}  \ottsym{,}  \ottmv{i}  \ottsym{)} }  +
   \Wparam{conv}{ \varsigma _{ \mathit{t} } }  \mathit{v} _{ \mathit{t} }  +
   \Wparam{conv}{ \varsigma _{ \mathit{t} } , p}  \mathit{v} _{ \textsf{parent} \, \ottsym{(}  \mathit{t}  \ottsym{)} }  +
   \bparam{conv}{ \varsigma _{ \mathit{t} } }
 \right) \label{eqn:tcl}
\end{equation}
where $\Wparam{conv}{ \varsigma _{ \mathit{t} } ,i} \in \RealMat{m}{n}$ is a weight
parameter which is a coefficient of the feature vector of the $i$-th
child, $\Wparam{conv}{ \varsigma _{ \mathit{t} } } \in \RealMat{m}{n}$ is for
$\mathit{t}$, $\Wparam{conv}{ \varsigma _{ \mathit{t} } ,p} \in \RealMat{m}{n}$ is for
the parent, $\bparam{conv}{ \varsigma _{ \mathit{t} } } \in \RealVec{m}$ is a
bias parameter for $ \varsigma _{ \mathit{t} } $, and
\Fact{conv} is an activation function.
Each parameter is shared between nodes with the same $ \varsigma _{ \mathit{t} } $.
If $\mathit{t}$ is the root node, then $ \mathit{v} _{ \textsf{parent} \, \ottsym{(}  \mathit{t}  \ottsym{)} } $ denotes the zero
vector.
We use multiple AST convolution layers to learn features of a node.

The update (\ref{eqn:tcl}) is inspired by tree-based convolution
proposed by \citet{Mou_2016_AAAI}, but there are a few differences.
First, our update rule involves the feature vector of a parent node to capture
features of the context where a node is used, whereas Mou et al.\ do not.
Second, Mou et al.\ regard an AST as a binary tree, which is possible, e.g., by
left-child right-sibling binary trees,\footnote{They did not clarify what binary
tree is considered, though.} which makes it possible to fix the number of weight
parameters for children to be only two.
This view is useful when one deals with ASTs where a node may have an arbitrary
number of children.
However, a different tree representation may affect a feature representation
learned by DNNs---especially, it may not preserve the locality of the original
AST representation.
Thus, instead of binary trees, we deal with ASTs as they are.
Fortunately, the syntax of propositions is defined rigorously and the number of
children of each node is fixed.
Hence, we can fix the number of learnable weight parameters for children:
two for each proposition constructor.

\paragraph{Aggregation layer}
An aggregation layer integrates features of nodes in an AST to a single vector.
\begin{definition}[Aggregation layer]
 Let $\mathit{t}$ be a node of an AST where nodes are augmented with $n$-dimensional vectors.
 Function $\textsf{Agg} \, \ottsym{(}  \mathit{t}  \ottsym{)}$ produces an $n$-dimensional vector from $\mathit{t}$
 as follows:
 \[
  \textsf{Agg} \, \ottsym{(}  \mathit{t}  \ottsym{)} = \Fact{agg} \left(
    \sum_{i} \Wparam{agg}{ \varsigma _{ \mathit{t} } ,i} \, \textsf{Agg} \, \ottsym{(}  \textsf{child} \, \ottsym{(}  \mathit{t}  \ottsym{,}  \ottmv{i}  \ottsym{)}  \ottsym{)} +
    \Wparam{agg}{ \varsigma _{ \mathit{t} } }  \mathit{v} _{ \mathit{t} }  +
    \bparam{agg}{ \varsigma _{ \mathit{t} } }
 \right)
 \]
 where $\Wparam{agg}{ \varsigma _{ \mathit{t} } }$ and $\Wparam{agg}{ \varsigma _{ \mathit{t} } ,i} \in
 \RealMat{n}{n}$ are weight parameters which are coefficients of vectors of
 $\mathit{t}$ and its $i$-th child, respectively, $\bparam{agg}{ \varsigma _{ \mathit{t} } } \in
 \RealVec{n}$ is a bias for $ \varsigma _{ \mathit{t} } $, and $\Fact{agg}$ is an
 activation function.
 Each parameter is shared between nodes with the same $ \varsigma _{ \mathit{t} } $.
\end{definition}

Another way to produce a single feature vector from an AST is a
max-pool~\cite{Mou_2016_AAAI}, which, for each dimension, takes the maximum
scalar value among all feature vectors in the AST.
While max-pools are used by usual convolutional neural
networks~\cite{Alex_2012_NIPS}, it is not clear that gathering only maximum
values captures features of the whole of the AST.
By contrast, an aggregation layer can be considered as ``fold'' on trees with
feature vectors, and we expect that $\textsf{Agg} \, \ottsym{(}  \mathit{t}  \ottsym{)}$ learns a feature
representation of the AST because it takes not only maximum values but also the
other elements of feature vectors of all nodes into account.
\TS{Indeed, we confirm that the aggregation layer outperforms the max-pool in
\sect{}.}

In what follows, we write $ \mathit{v} _{ \ottnt{P} } $ for the feature vector of $\ottnt{P}$ that
is achieved by applying $ \textsf{Enc} $, multiple AST convolution layers, and an
aggregation layer sequentially.

\subsubsection{Path encoder}
\label{sec:model-deep-decoder}
To achieve good performance, we have to know what assumptions are
available at the position for which an inference rule is estimated.
For example, if a variable of type $\mathit{a} \,  \mathord{\rightarrow}  \, \mathit{b}$ can be referred to, we expect the
variable to be useful to prove $\mathit{b}$.
\KS{Rephrase this sentense.}
We can access to information of assumptions via proposition $\ottnt{P}$ and path
$\rho$.
The proposition-to-proof model thus extracts features of assumptions from
the feature vector $ \mathit{v} _{ \ottnt{P} } $ of $\ottnt{P}$ along the given $\rho$.
\TS{Have we discussed $\beta\eta$ normal forms are desired?
We also expect paths to be useful to understand a context where an inference
rule is estimated; it is important when we are interested in synthesis of proofs
in a $\beta\eta$ normal form.
For example, given a partially constructed proof $ \left[ \, \right] _{  0  }  \,  \left[ \, \right] _{  1  } $, we hope that the
model does not estimate that $ \left[ \, \right] _{  0  } $ should be filled with an lambda
abstraction because $\ottsym{(}  \lambda  \mathit{x}  \ottsym{.}  \ottnt{M}  \ottsym{)} \, \ottnt{N}$ is $\beta$-reducible.
}

\begin{definition}[Extraction]
 \label{defn:extract}
 $\textsf{Extract} \, \ottsym{(}  \rho  \ottsym{,}  \mathit{v}  \ottsym{)}$ extracts features in the position to which $\rho$
 points from $\mathit{v}$.
 \[\begin{array}{lll}
  \textsf{Extract} \, \ottsym{(}   \langle \rangle   \ottsym{,}  \mathit{v}  \ottsym{)} &=& \Wparam{ext}{} \, \mathit{v} + \bparam{ext}{} \\
  \textsf{Extract} \, \ottsym{(}    \langle  \ottsym{(}  \ottnt{C}  \ottsym{,}  \ottmv{i}  \ottsym{)} ,  \rho  \rangle    \ottsym{,}  \mathit{v}  \ottsym{)} &=& \textsf{Extract} \, \ottsym{(}  \rho  \ottsym{,}  \mathit{v}'  \ottsym{)}
   \quad \text{ where }
     \mathit{v}' = \Fact{ext}\left( \Wparam{ext}{\ottnt{C}, i}\, \mathit{v} + \bparam{ext}{\ottnt{C}, i} \right)
   \end{array}
 \]
 where $\Wparam{ext}{}$, $\Wparam{ext}{\ottnt{C}, i} \in \RealMat{n}{n}$
 and $\bparam{ext}{}$, $\bparam{ext}{\ottnt{C}, i} \in \RealVec{n}$ are
 learnable parameters and $\Fact{ext}$ is an activation function.
 $ \langle  \ottsym{(}  \ottnt{C}  \ottsym{,}  \ottmv{i}  \ottsym{)} ,  \rho  \rangle $ is the addition of $\ottsym{(}  \ottnt{C}  \ottsym{,}  \ottmv{i}  \ottsym{)}$ to path $\rho$
 at the beginning.
\end{definition}

\begin{figure}[t]
 \includegraphics[width=\textwidth]{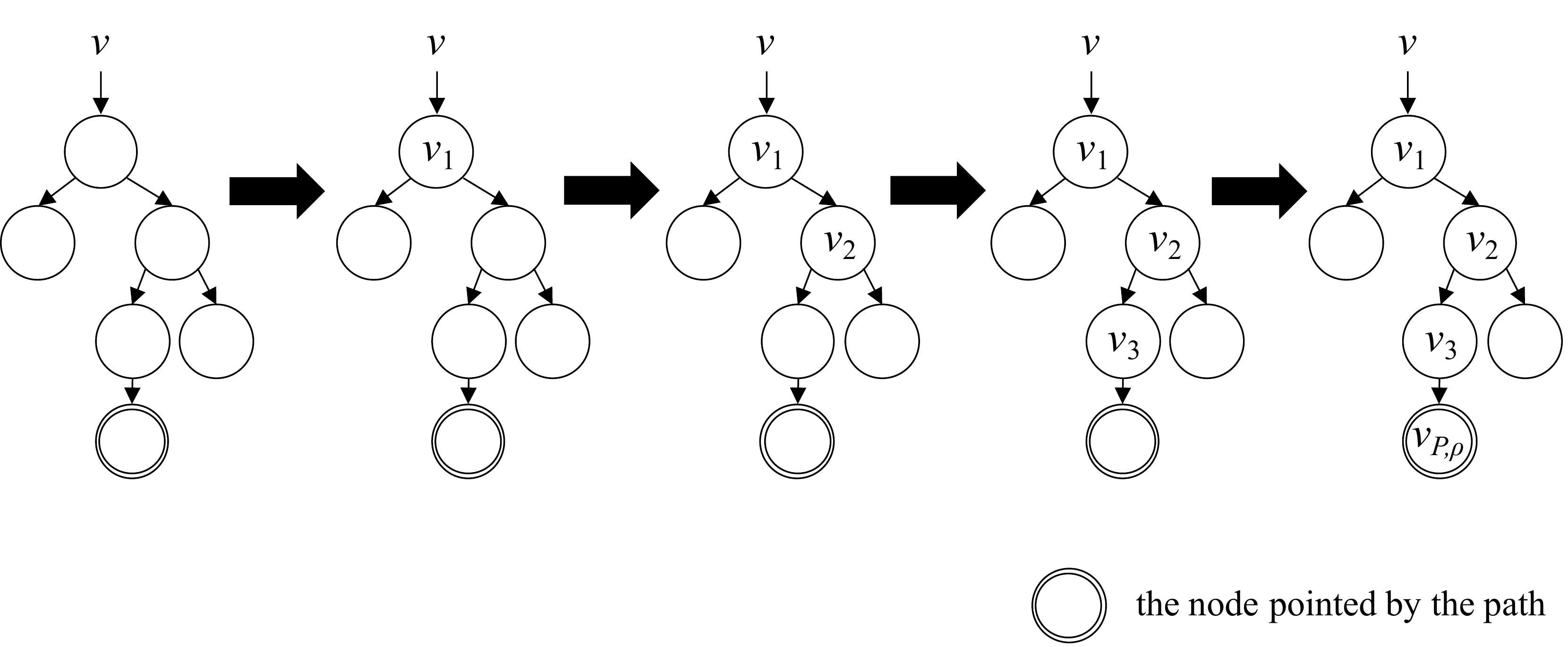}
 \caption{A running example of extraction.}
 \label{fig:extract}
\end{figure}
\fig{extract} illustrates the process of computing $\textsf{Extract} \, \ottsym{(}  \rho  \ottsym{,}  \mathit{v}  \ottsym{)}$,
where features are extracted along the path.
We write $ \mathit{v} _{ \ottnt{P}  \ottsym{,}  \rho } $ for $\textsf{Extract} \, \ottsym{(}  \rho  \ottsym{,}   \mathit{v} _{ \ottnt{P} }   \ottsym{)}$.
The weight parameters in \defn{extract} have a role of extracting features
necessary to capture assumptions from $ \mathit{v} _{ \ottnt{P} } $.
The biases are expected to capture information of the context around the node to
which the path points.

\subsubsection{Classification}
We estimate what proof inference rule is most likely to be applied by using two
feature vectors $ \mathit{v} _{ \ottnt{P}  \ottsym{,}  \rho } $, the extracted features from $\ottnt{P}$ along
$\rho$, and $ \mathit{v} _{ \ottnt{Q} } $, the features of proof obligation $\ottnt{Q}$.
For that, as usual, we concatenate $ \mathit{v} _{ \ottnt{P}  \ottsym{,}  \rho } $ and $ \mathit{v} _{ \ottnt{Q} } $ and apply
multiple fully connected layers to the concatenation result so that the number
of dimensions of the final output $ \mathit{v} _{  \mathrm{o}  } $ is equal to that of proof
inference rules, that is, eight.
Using $ \mathit{v} _{  \mathrm{o}  } $, we approximate the likelihood of a proof inference rule
$\mathit{r}$ being applied by softmax.
For vector $\mathit{v} \in \RealVec{n}$, we write $ \mathit{v}  [  \ottmv{i}  ] $ for the real number
of the $i$-th dimension of $\mathit{v}$.
Let $ \ottmv{n} _{ \mathit{r} }  \in \{1,...,8\}$ be an index corresponding to proof inference rule
$\mathit{r}$ in $ \mathit{v} _{  \mathrm{o}  } $.
Then, the approximation probability
$ \mathit{p} ^*   \ottsym{(}   \boldsymbol{r}  =  \mathit{r}   \mid   \boldsymbol{P}  =  \ottnt{P}   \ottsym{,}   \boldsymbol{\rho}  =  \rho   \ottsym{,}   \boldsymbol{Q}  =  \ottnt{Q}   \ottsym{)}$ is calculated by:
\[
   \frac{exp(  \mathit{v} _{  \mathrm{o}  }   [    \ottmv{n} _{ \mathit{r} }    ] )}{\sum_{j=1}^{8} exp(  \mathit{v} _{  \mathrm{o}  }   [  \ottmv{j}  ] )}
\]

\section{Experiments}
\label{sec:exp}

This section reports the performance of our proposition-to-proof model and the
proof synthesis procedure with it.
We train the model on a dataset that contains pairs of a proposition and its
proof by supervised learning.
After explaining the detailed architecture of our model (\sect{exp-config}), we
detail creation of the dataset (\sect{exp-dataset}).
We evaluate the trained model on the basis of accuracy, that is, we check, given
a proposition, a partially constructed proof, and a hole from a validation
dataset, how accurately the model estimates the inference rule to be applied at
the hole; we also conduct an in-depth analysis of the model to confirm how
influential depths of hole positions are on the accuracy (\sect{exp-acc}).
Finally, we evaluate the proof synthesis procedure given in
\sect{proof-synthesis} (\sect{exp-proof}).

We implemented the procedure \textsc{ProofSynthesize} and our model on
Python 3 (version 3.6.3) with the deep learning framework
Chainer~\cite{tokui2015chainer} (version 2.1.0).
We use the Haskell interpreter GHCi (version 8.0.1) for a type checker
in \textsc{ProofSynthesize}.
All experiments are conducted on a machine equipped with 12 CPU cores (Intel
i7-6850K 3.60GHz), 32 GB RAM, and NVIDIA GPUs (Quadro P6000).

\subsection{Network configuration}
\label{sec:exp-config}

\begin{figure}[t]
 \includegraphics[width=\textwidth]{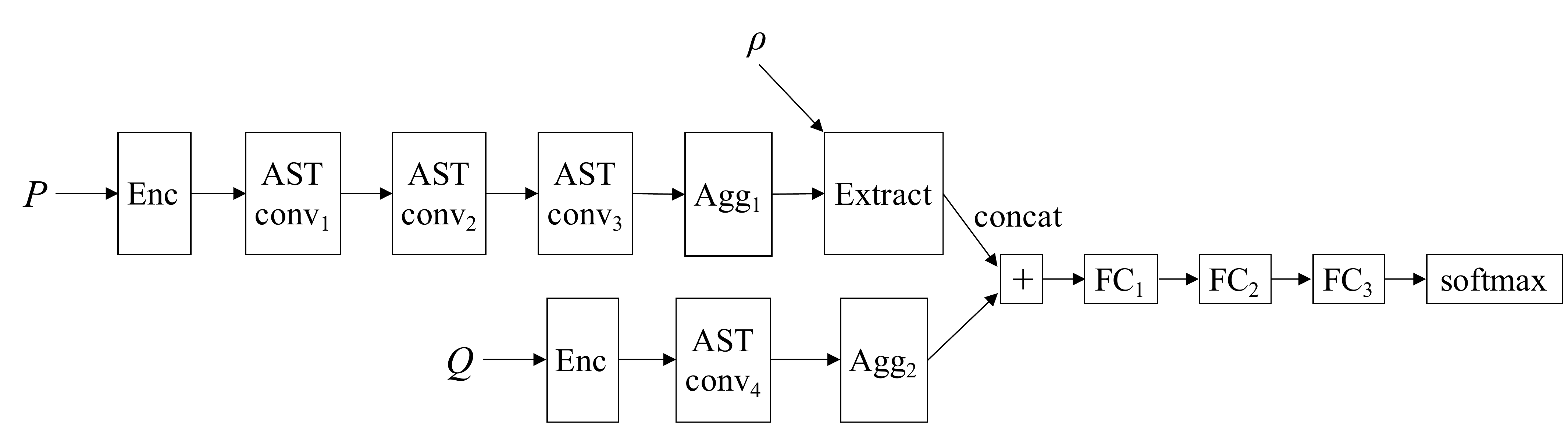}
 \caption{The architecture of our DNN model. $\ottnt{P}$ is a proposition to be
 proven, $\rho$ is a path specifying the hole to be filled, and $\ottnt{Q}$ is
 a proof obligation to be discharged at the hole.}
 \label{fig:arch}
\end{figure}
\begin{table}[t]
 {\tabulinesep=0.7mm
 \begin{tabu}{l@{\qquad}lc}
  Layer        & Learnable parameters & Number of dimensions of output vectors \\ \hline \hline
  \multirow{2}{*}{AST conv$_1$} &
      $\Wparam{conv}{\varsigma}, \Wparam{conv}{\varsigma,i}, \Wparam{conv}{\varsigma,p} \in \RealMat{200}{4}$ &
      \multirow{2}{*}{200} \\ &
      $\bparam{conv}{\varsigma} \in \RealVec{200}$
      \\ \hline
  \multirow{2}{*}{AST conv$_2$} &
      $\Wparam{conv}{\varsigma}, \Wparam{conv}{\varsigma,i}, \Wparam{conv}{\varsigma,p} \in \RealMat{500}{200}$ &
      \multirow{2}{*}{500} \\ &
      $\bparam{conv}{\varsigma} \in \RealVec{500}$
      \\ \hline
  \multirow{2}{*}{AST conv$_3$} &
      $\Wparam{conv}{\varsigma}, \Wparam{conv}{\varsigma,i}, \Wparam{conv}{\varsigma,p} \in \RealMat{1000}{500}$ &
      \multirow{2}{*}{1000} \\ &
      $\bparam{conv}{\varsigma} \in \RealVec{1000}$
      \\ \hline
  $ \textsf{Agg} _1$ &
      $\Wparam{agg}{\varsigma}, \Wparam{agg}{\varsigma,i} \in \RealMat{1000}{1000}$ \ $\bparam{agg}{\varsigma} \in \RealVec{1000}$ &
      1000 \\ \hline
  \multirow{2}{*}{$ \textsf{Extract} $} &
      $\Wparam{ext}{}$, $\Wparam{ext}{\ottnt{C}, i} \in \RealMat{1000}{1000}$ &
      \multirow{2}{*}{1000} \\ &
      $\bparam{ext}{}, \bparam{ext}{\ottnt{C}, i} \in \RealVec{1000}$
      \\ \hline
  \multirow{2}{*}{AST conv$_4$} &
      $\Wparam{conv}{\varsigma}, \Wparam{conv}{\varsigma,i}, \Wparam{conv}{\varsigma,p} \in \RealMat{16}{4}$ &
      \multirow{2}{*}{16} \\ &
      $\bparam{conv}{\varsigma} \in \RealVec{16}$
      \\ \hline
  $ \textsf{Agg} _2$ &
      $\Wparam{agg}{\varsigma}, \Wparam{agg}{\varsigma,i} \in \RealMat{16}{16}$ \ $\bparam{agg}{\varsigma} \in \RealVec{16}$ &
      16 \\ \hline
  $\mathrm{FC}_1$ & $\Wparam{fc}{} \in \RealMat{1016}{1016}$ \hspace{3ex} $\bparam{fc}{} \in \RealVec{1016}$ & 1016 \\ \hline
  $\mathrm{FC}_2$ & $\Wparam{fc}{} \in \RealMat{1016}{1016}$ \hspace{3ex} $\bparam{fc}{} \in \RealVec{1016}$ & 1016 \\ \hline
  $\mathrm{FC}_3$ & $\Wparam{fc}{} \in \RealMat{8}{1016}$    \hspace{5.7ex} $\bparam{fc}{} \in \RealVec{8}$    & 8    \\ \hline
 \end{tabu}} \\[1ex] \mbox{}
 \caption{Learnable parameters and the number of dimensions of vectors in the
 output for each layer. $\varsigma$ is a class of a node in a proposition AST.
 }
 \label{tbl:arch}
\end{table}
\fig{arch} shows the architecture of our
proposition-to-proof model in the experiments.
We use three AST convolution layers to encode proposition $\ottnt{P}$ to be proven
and one for proof obligation $\ottnt{Q}$.
The concatenation result of $ \mathit{v} _{ \ottnt{P}  \ottsym{,}  \rho } $ from $ \textsf{Extract} $ and
$ \mathit{v} _{ \ottnt{Q} } $ from $ \textsf{Agg} _2$ is fed to three fully connected layers.
The detailed specification of each layer is shown in \tbl{arch}.
We use a rectified linear unit (ReLU)~\cite{DBLP:journals/jmlr/GlorotBB11} as
activation functions throughout the architecture.

\subsection{Dataset}
\label{sec:exp-dataset}
The power of deep learning rests on datasets used to train DNN models.
In this work, we need a dataset of pairs of a proposition and its proof.
We make dataset {\datasetall} by generating small proofs exhaustively
and large proofs at random.

\begin{algorithm}[t]
 \caption{Small proof generation}
 \label{alg:gen-small-proof}
 \begin{algorithmic}[1]
  \Procedure{SmallProofGen}{$s$}
  \State Initialize {\dataset} with the empty set and
                    {\termset} with the empty queue
  \State Push $ \left[ \, \right] $ to {\termset}
  \While{{\termset} is not empty}
    \State Pop $\ottnt{M}$ from {\termset}
    \State Let $ \left[ \, \right] _{ \ottmv{i} } $ be the leftmost hole in $\ottnt{M}$
    \For{each $\ottnt{C} \,  \in  \, \textsf{Cnstr} \, \ottsym{(}  \ottnt{M}  \ottsym{,}   \left[ \, \right] _{ \ottmv{i} }   \ottsym{)}$ such that $ \ottnt{M}  [  \ottnt{C}  ]_{ \ottmv{i} } $ is a $\beta\eta$ normal form, \par \qquad\qquad $\textsf{size} \, \ottsym{(}   \ottnt{M}  [  \ottnt{C}  ]_{ \ottmv{i} }   \ottsym{)} \le s$, and $ \emptyset   \vdash   \ottnt{M}  [  \ottnt{C}  ]_{ \ottmv{i} }  \,  \mathrel{:}  \, \ottnt{P}$ for principal $\ottnt{P}$}
      \If{$ \ottnt{M}  [  \ottnt{C}  ]_{ \ottmv{i} } $ has a hole}
        \State Push $ \ottnt{M}  [  \ottnt{C}  ]_{ \ottmv{i} } $ to {\termset}
     \ElsIf{{\dataset} contains $(\ottnt{P},\ottnt{N})$ such that $\textsf{size} \, \ottsym{(}  \ottnt{M}  \ottsym{)} < \textsf{size} \, \ottsym{(}  \ottnt{N}  \ottsym{)}$} \label{alg:gen-small-proof-contained}
        \State {\dataset} $\leftarrow ({\dataset} \mathop{\backslash} \{ (\ottnt{P},\ottnt{N})) \} \cup \{ (\ottnt{P}, \ottnt{M}  [  \ottnt{C}  ]_{ \ottmv{i} } ) \}$ \label{alg:gen-small-proof-contained-end}
     \ElsIf{{\dataset} does not contain $(\ottnt{P},\ottnt{N})$ for any $\ottnt{N}$}
        \State {\dataset} $\leftarrow {\dataset} \cup \{ (\ottnt{P}, \ottnt{M}  [  \ottnt{C}  ]_{ \ottmv{i} } ) \}$
      \EndIf
    \EndFor
  \EndWhile
  \State \Return {\dataset}
  \EndProcedure
 \end{algorithmic}
\end{algorithm}
Small proofs are generated by \proc{gen-small-proof}, which produces a set
${\dataset}$ of pairs of a proposition and its proof the size of which is equal
to or less than $s$.
\proc{gen-small-proof} generates proofs by filling the leftmost holes $ \left[ \, \right] _{ \ottmv{i} } $
in terms $\ottnt{M}$ of queue {\termset}.
If proposition $\ottnt{P}$ of a generated proof $\ottnt{M}$ is already included jointly
with $\ottnt{N}$ in {\dataset}, we choose the proof the size of which is smaller
(Lines~\ref{alg:gen-small-proof-contained}--\ref{alg:gen-small-proof-contained-end})
in order to decrease the number of estimations performed by
\textsc{ProofSynthesize}---the error by approximation becomes larger as more
estimations are performed.
We call $\ottnt{P}$ principal when, for any $\ottnt{Q}$ such that $ \emptyset   \vdash  \ottnt{M} \,  \mathrel{:}  \, \ottnt{Q}$,
there exists some map from propositional variables to propositions such that
$f(\ottnt{P}) = \ottnt{Q}$~\cite{Milner_1978_JCSS}; well-typed terms in the simply typed
lambda calculus have principal types.
Following \citet{DBLP:journals/corr/SekiyamaIS17}, we have constructed
only $\beta\eta$ normal forms.
%
%
The dataset {\datasetall} that we use in this work includes a dataset produced
by \textsc{SmallProofGen(9)}.

Generating large proofs is not so easy due to the huge space to be searched.
We generate a large proof efficiently, as follows.
Suppose that a lower bound $l$ and an upper bound $u$ of the size of a proof
generated are given and let $\ottnt{M}$ be a $\beta\eta$ normal proof partially
constructed so far.
We start with $\ottnt{M} \, \ottsym{=} \, \left[ \, \right]$.
We gradually fill holes in $\ottnt{M}$ with term constructors chosen randomly and
keep $\ottnt{M}$ to be the partial proof produced last.
If $\ottnt{M}$ becomes a complete proof with a smaller size than $l$, we restart the
proof generation from the beginning with $\ottnt{M} \, \ottsym{=} \, \left[ \, \right]$.
If the size of $\ottnt{M}$ becomes larger than $l$, we preferentially
choose variables as term constructors substituted for holes to finish the proof
generation as soon as possible.
If $\ottnt{M}$ becomes a complete proof with size $s$ such that $l \le s \le u$, we
produce $\ottnt{M}$ as the result.
If the size of $\ottnt{M}$ becomes large than $u$, we restart the proof
generation.
This approach may appear rather ad-hoc, but we could generate many large proofs
by it.
For $(l,u) \in \{ (10,30), (20, 40), (30, 50) \}$, we generate 15000, 15000, and
10000 proofs, respectively.

\begin{table}[t]
 \begin{tabu}{crr}
  \multicolumn{1}{c}{Size} & \multicolumn{1}{c}{Number of proofs} \\ \hline \hline
  1--10  & 136877 \qquad\quad \\
  11--20 & 14885 \qquad\quad \\
  21--30 & 7910 \qquad\quad \\
  31--40 & 5848 \qquad\quad \\
  41--50 & 2224 \qquad\quad \\ \hline
  1--50  & 167744 \qquad\quad
 \end{tabu}
 \caption{The number of proofs per size in {\datasetall}}
 \label{tbl:proof-dataset}
\end{table}
The dataset {\datasetall} contains proofs shown in \tbl{proof-dataset}.
Since our DNN model feeds a proposition $\ottnt{P}$, a path $\rho$, and a proof
obligation $\ottnt{Q}$ to be discharged at the hole specified by $\rho$ and estimates an inference rule
$\mathit{r}$ that should be applied at the hole, we make quadruples
$(\ottnt{P}, \ottnt{Q}, \rho, \mathit{r})$ from {\datasetall} and split them into
training dataset {\datasett} and validation dataset {\datasetv}.
{\datasett} contains 90\% of quadruples generated from {\datasetall} (1731998
quadruples) and {\datasetv} does the remaining 10\% (193108 ones).
\begin{table}[t]
 \begin{tabu}{cr}
  Inference rule            & Number of training data \\ \hline \hline
  \ottdrulename{Var}        & 453655 \quad\quad\qquad \\
  \ottdrulename{Abs}        & 473338 \quad\quad\qquad \\
  \ottdrulename{App}        & 29621 \quad\quad\qquad \\
  \ottdrulename{Pair}       & 172268 \quad\quad\qquad \\
  \ottdrulename{CasePair}   & 27272 \quad\quad\qquad \\
  \ottdrulename{Left}       & 269613 \quad\quad\qquad \\
  \ottdrulename{Right}      & 269480 \quad\quad\qquad \\
  \ottdrulename{CaseEither} & 36750 \quad\quad\qquad \\
 \end{tabu}
 \caption{The number of training data for each inference rule.}
 \label{tbl:training-dataset}
\end{table}
\tbl{training-dataset} shows the number of training data in {\datasett} for
each inference rule.

\subsection{Training}
\label{sec:exp-train}
We train the proposition-to-proof model with the architecture given in
\sect{exp-config} on dataset {\datasett} by stochastic gradient descent with a
mini-batch size of 1000 for 20 epochs.\footnote{Epoch is the unit that means how
many times the dataset is scanned during the training.}
Weights in each layer of the model are initialized by the values
independently drawn from the Gaussian distribution with mean $0$ and
standard deviation $\sqrt{\frac{1}{n}}$ where $n$ is the number of
dimensions of vectors in the input to the layer.
The biases are initialized with 0.
We use the softmax cross entropy as the loss function.
As an optimizer, we use Adam~\cite{DBLP:journals/corr/KingmaB14} with parameters $\alpha = 0.001$, $\beta_1
= 0.9$, $\beta_2 = 0.999$, and $\epsilon = 10^{-8}$.
We lower $\alpha$, which controls the learning rate, by 10 times when the
training converges.
We regularize our model by a weight decay with penalty rate $\lambda = 0.0001$.

\subsection{Evaluation}
\subsubsection{Accuracy}
\label{sec:exp-acc}

\begin{table}[t]
 \small
 \begin{tabu}{c|cccccccccc}
  Depth  & \#     & All   & \ottdrulename{Var} & \ottdrulename{Abs} & \ottdrulename{App} & \ottdrulename{Pair} & \ottdrulename{CasePair} & \ottdrulename{Left} & \ottdrulename{Right} & \ottdrulename{CaseEither} \\ \hline \hline
  1      & 16774  & 100.0 & N/A                & 100.0              & N/A                & 100.0               & N/A                     & 100.0               & 100.0                & N/A                       \\
  2      & 18427  & 99.32 & N/A                & 99.77              & 94.17              & 99.69               & 89.31                   & 99.61               & 99.76                & 84.68                     \\
  3      & 21932  & 98.34 & 98.38              & 99.61              & 92.54              & 99.62               & 76.09                   & 99.21               & 98.97                & 79.19                     \\
  4      & 25250  & 97.72 & 97.46              & 99.50              & 89.13              & 99.52               & 72.97                   & 98.81               & 98.78                & 83.41                     \\
  5      & 27262  & 96.92 & 97.36              & 99.30              & 90.95              & 98.82               & 62.87                   & 97.50               & 97.75                & 82.08                     \\
  6      & 26107  & 96.63 & 98.53              & 99.10              & 85.87              & 98.63               & 53.42                   & 97.12               & 96.25                & 70.96                     \\
  7      & 20719  & 96.68 & 98.76              & 98.82              & 80.46              & 97.39               & 50.52                   & 97.93               & 97.31                & 48.11                     \\
  8      & 13466  & 95.27 & 98.45              & 98.35              & 35.48              & 95.33               & 37.05                   & 96.59               & 96.58                & 38.05                     \\
  9      & 7616   & 92.57 & 97.90              & 96.16              & 30.51              & 92.46               & 34.10                   & 93.88               & 95.86                & 33.00                     \\
  10     & 4639   & 90.54 & 96.62              & 97.19              & 37.66              & 91.11               & 24.80                   & 93.18               & 92.44                & 32.87                     \\
  11     & 3478   & 90.80 & 96.88              & 97.04              & 20.97              & 93.60               & 26.00                   & 92.57               & 93.70                & 29.41                     \\
  12     & 2460   & 89.59 & 96.51              & 97.30              & 21.67              & 91.45               & 15.39                   & 91.41               & 93.07                & 26.92                     \\
  13     & 1735   & 88.59 & 96.26              & 95.50              & 23.91              & 83.15               & 16.67                   & 94.24               & 95.35                & 16.98                     \\
  14     & 1221   & 90.17 & 98.08              & 97.06              & 12.00              & 89.39               & 21.88                   & 93.97               & 89.92                & 17.14                     \\
  15     & 790    & 90.13 & 98.00              & 97.89              & 29.41              & 97.62               & 6.67                    & 87.14               & 88.75                & 12.00                     \\
  16--20 & 1158   & 90.50 & 97.91              & 95.75              & 19.05              & 91.11               & 8.33                    & 95.05               & 86.79                & 3.70                      \\
  21--26 & 74     & 90.54 & 97.50              & 85.71              & 0.00               & 100.0               & 0.00                    & 75.00               & 100.0                & N/A                       \\ \hline
  1--26  & 193108 & 96.79 & 98.03              & 99.27              & 78.21              & 98.50               & 57.25                   & 98.05               & 97.95                & 67.34                     \\
 \end{tabu}
 \caption{Validation accuracy of the trained model for each inference rule per
 depth. The column ``\#'' shows the number of validation data and ``All'' does
 the accuracy for all inference rules. ``N/A'' means that there are no
 validation data.}  \label{tbl:acc}
\end{table}

\tbl{acc} shows the accuracy of the trained model on the validation dataset
{\datasetv}.
%
%
The bottom row in the table reports
the summarized accuracy and presents that the trained model achieves total
accuracy 96.79\%.
Looking at results per inference rule, we achieve the very high accuracy for
\ottdrulename{Var}, \ottdrulename{Abs}, \ottdrulename{Pair},
\ottdrulename{Left}, and \ottdrulename{Right}.
It is interesting that the train model chooses either of \ottdrulename{Left} or
\ottdrulename{Right} appropriately according to problem instances.
It means that, given proposition $\ottnt{P} \,  \mathord{+}  \, \ottnt{Q}$, the proof synthesis procedure
with this trained model can select whichever of $\ottnt{P}$ and $\ottnt{Q}$ should be
proven with high probability.
The accuracy for \ottdrulename{App}, \ottdrulename{CasePair}, and
\ottdrulename{CaseEither} is not so bad, but the
estimation of these rules is more difficult than that of other rules.
This may be due to the training dataset.
As shown in \tbl{training-dataset}, the numbers of training data for
\ottdrulename{App}, \ottdrulename{CasePair}, and \ottdrulename{CaseEither} are
much smaller than those of other rules.
Since the model is trained so that inference rules that often occur in the
training dataset are more likely to be estimated in order to minimize the loss,
the trained model may prefer to choose inference rules other than
\ottdrulename{App}, \ottdrulename{CasePair}, and \ottdrulename{CaseEither}.
Furthermore, it may be possible that the training data for those rules are
insufficient to learn feature representation of the likelihood of them being
applied.
In either case, data augmentation would be useful, though we need to establish
effective augmentation of proofs.

Our model is supposed to access the assumptions via the $\ottnt{P}$ and $\rho$.
Since $\rho$ becomes larger as the position of the hole does deeper, the
depth of the hole is expected to affect the performance of the model.
We thus investigate the accuracy of the trained model for each depth of holes in
the validation data, which is shown in \tbl{acc}.
Seeing the column ``All'', we can find that the accuracy at a greater depth
tends to be lower.
The accuracy of \ottdrulename{Abs}, \ottdrulename{Pair}, \ottdrulename{Left},
and \ottdrulename{Right} is still high even if holes are at deep positions.
We consider that this is because, rather than assumptions, proof obligations play
an important role to choose those inference rules.
By contrast, the accuracy of \ottdrulename{App}, \ottdrulename{CasePair}, and
\ottdrulename{CaseEither} is not high, especially, when holes are at very deep
positions.
Since these rules need information about assumptions to judge whether they
should be applied, their accuracy may be improved by representing features of
assumptions better.
The accuracy of \ottdrulename{Var} is very high at any depth, though whether we
can apply \ottdrulename{Var} should depend on assumptions.
This may be due to the large number of training data for \ottdrulename{Var}
(\tbl{training-dataset}), which may make it possible to learn feature
representation of assumptions only for \ottdrulename{Var}.

Finally, we confirm the power of explicit use of proof obligations.
To this end, we train a model that does \emph{not} use the feature vector of a
proof obligation; we call such a model obligation-free.
The architecture of the obligation-free model is the same as \fig{arch} except
that it does not refer to the feature vector of proof obligation $\ottnt{Q}$.
We train the obligation-free model in the same way as \sect{exp-train}.
\begin{table}[t]
 \small
 \begin{tabu}{c|cccccccccc}
  Depth  & \#     & All   & \ottdrulename{Var} & \ottdrulename{Abs} & \ottdrulename{App} & \ottdrulename{Pair} & \ottdrulename{CasePair} & \ottdrulename{Left} & \ottdrulename{Right} & \ottdrulename{CaseEither} \\ \hline \hline
  1      & 16774  & 99.99 & N/A                & 100.0              & N/A                & 100.0               & N/A                     & 99.97               & 99.97                & N/A                       \\
  2      & 18427  & 99.07 & N/A                & 99.68              & 94.17              & 99.15               & 86.16                   & 99.36               & 99.55                & 85.96                     \\
  3      & 21932  & 97.24 & 97.73              & 98.92              & 88.06              & 97.99               & 76.09                   & 98.12               & 96.93                & 82.14                     \\
  4      & 25250  & 95.17 & 96.28              & 97.49              & 87.11              & 96.20               & 63.29                   & 96.33               & 95.21                & 81.61                     \\
  5      & 27262  & 92.19 & 96.05              & 95.15              & 85.56              & 91.99               & 53.22                   & 91.54               & 89.97                & 76.63                     \\
  6      & 26107  & 89.12 & 96.71              & 91.77              & 77.81              & 82.93               & 39.21                   & 86.47               & 83.08                & 59.35                     \\
  7      & 20719  & 85.18 & 97.18              & 86.90              & 64.52              & 60.79               & 26.12                   & 76.89               & 74.80                & 18.21                     \\
  8      & 13466  & 78.15 & 95.93              & 77.79              & 13.71              & 28.31               & 13.84                   & 59.69               & 55.87                & 9.74                      \\
  9      & 7616   & 66.19 & 93.62              & 66.67              & 7.63               & 15.28               & 4.05                    & 28.79               & 20.86                & 5.08                      \\
  10     & 4639   & 56.33 & 87.81              & 67.95              & 7.79               & 10.28               & 6.40                    & 19.89               & 20.48                & 0.70                      \\
  11     & 3478   & 56.67 & 88.19              & 69.03              & 8.07               & 9.20                & 1.00                    & 14.59               & 13.85                & 2.35                      \\
  12     & 2460   & 57.24 & 89.43              & 73.28              & 6.67               & 7.24                & 3.85                    & 11.34               & 12.04                & 2.56                      \\
  13     & 1735   & 57.06 & 86.18              & 72.97              & 4.35               & 9.95                & 10.42                   & 9.95                & 13.95                & 1.89                      \\
  14     & 1221   & 59.46 & 86.38              & 75.98              & 4.00               & 10.61               & 3.13                    & 10.35               & 7.56                 & 5.71                      \\
  15     & 790    & 58.48 & 85.71              & 72.54              & 5.88               & 4.76                & 0.00                    & 5.71                & 4.76                 & 8.00                      \\
  16--20 & 1158   & 61.66 & 86.50              & 74.06              & 9.52               & 11.11               & 8.33                    & 3.96                & 5.66                 & 0.00                      \\
  21--26 & 74     & 64.86 & 85.00              & 85.71              & 0.00               & 0.00                & 0.00                    & 12.5                & 11.11                & N/A                       \\ \hline
  1--26  & 193108 & 88.52 & 94.06              & 94.06              & 69.40              & 83.90               & 44.42                   & 85.69               & 84.37                & 57.41                     \\
 \end{tabu}
 \caption{Validation accuracy of the trained obligation-free model.}
 \label{tbl:acc-obl-free}
\end{table}
The validation result of the trained obligation-free model is shown in
\tbl{acc-obl-free}.
Compared with \tbl{acc}, the accuracy of the obligation-free model is
lower than that of the proposition-to-proof model for all inference
rules, especially, at great depth.
The use of proof obligations thus improves the performance of the DNN model.

\subsubsection{Proof synthesis}
\label{sec:exp-proof}

This section evaluates \textsc{ProofSynthesize} (\proc{proof}) with the trained
proposition-to-proof model.
We make two test datasets for evaluation by choosing 500 propositions from
{\datasetv} respectively.
One dataset {\datasetsmall} consists of propositions that are generated by
\textsc{SmallProofGen(9)}, that is, the sizes of their proofs can be equal to or
lower than 9.
The other dataset {\datasetlarge} includes propositions that are generated at
random so that the sizes of their proofs are larger than 9.
We abort the proof synthesis if a proof is not generated within three
minutes.
We use the principal proposition for a proof obligation that is required by
\textsc{ProofSynthesize}.


We compare our procedure with an existing method of APS with deep learning by
\citet{DBLP:journals/corr/SekiyamaIS17}.
They view proof generation as a translation task from a proposition language to
a proof language and apply a so-called sequence-to-sequence model~\cite{Sutskever_2014_NIPS}, a
popular DNN model in machine translation, in order to produce a token sequence
expected to be a proof from a token sequence of a proposition.
They find that, though the response from the DNN model may not be a proof of the
proposition, the response is often ``close'' to a correct proof and, based on
this observation, propose a proof synthesis procedure that uses the response
from the DNN model as a guide of proof search.
We train the sequence-to-sequence model on {\datasett} for 200 epochs in the
same way as Sekiyama et al.\ and apply their proof synthesis procedure to
propositions in {\datasetsmall} and {\datasetlarge}.

\begin{table}[t]
 \tabulinesep=1.2mm
 \begin{tabu}{llcc} \hline
                                                          && \textsc{ProofSynthesize} & \citet{DBLP:journals/corr/SekiyamaIS17} \\ \hline
  \multirow{2}{*}{\datasetsmall} & Number of successes     & 500                     & 500 \\
                                 & Average time in success & 0.45                    & 1.85 \\ \hline
  \multirow{2}{*}{\datasetlarge} & Number of successes     & 466                     & 157 \\
                                 & Average time in success & 4.56                    & 29.03 \\ \hline
 \end{tabu}
 \caption{The evaluation result of the proof synthesis procedures: number of
 propositions that succeed in generation of proofs and average of elapsed times
 of the generation.}  \label{tbl:exp-proof}
\end{table}
\tbl{exp-proof} shows the number of propositions that succeed in generation of
proofs by each procedure and the average of elapsed times taken by the procedure
when proofs are generated successfully (the unit is second).
Both procedures succeed in generating proofs for all propositions in
{\datasetsmall}, which indicates that they work well, at least, for propositions
that have small proofs.
As for {\datasetlarge}, \textsc{ProofSynthesize} successfully generates proofs
for 93.2\% of propositions in {\datasetlarge}, while the procedure of Sekiyama
et al.\ does for only 31.4\%.
Since \textsc{ProofSynthesize} calculates the likelihood of a proof being a
correct one by the joint probability of inference rules in the proof, we can
generate a correct proof even in a case that the likelihoods of a few instances
of inference rules in the correct proof are estimated to be low, if the
likelihoods of other instances are to be high.
By contrast, the procedure of Sekiyama et al.\ uses only a single term as a
guide, so it is hard to recover the mistake of the estimation by the DNN model.
This would also lead to a difference of elapsed times taken by two proof
synthesis procedures---the procedure of Sekiyama et al.\ takes four times and
six times as long as \textsc{ProofSynthesize} for propositions in {\datasetsmall}
and {\datasetlarge}, respectively.
\section{Related work}
\label{sec:relatedWork}

\subsection{Automated theorem proving with deep learning}
Application of deep learning to ATP is becoming in trend recently.
%
Roughly speaking, there have been two research directions for ATP with deep
learning: \emph{enhancing existing solvers with deep learning} and
\emph{implementing ATP procedures using deep learning}.
We discuss these two lines of work in the following.


\subsubsection{Enhancing existing provers}
Existing automated theorem provers rely on many heuristics.  Applying
deep learning to improve these hand-crafted heuristics, aiming at
enhancing them, is an interesting direction.
%
\emph{Premise selection}, a task to select premises needed to prove a
given conjecture, is an important heuristic to narrow the search space
of proofs.
\citet{Irving_NIPS_2016} show the possibility of the application of
deep learning to this area using various DNN models to encode premises
and a conjecture to be proven in first-order logic.
\citet{Kaliszyk_arxiv_2017} make a dataset in the HOL Light theorem
prover~\cite{DBLP:conf/tphol/Harrison09a} for several tasks, including
premise selection, related to ATP.
\citet{Wang_2017_NIPS} tackles the premise selection problem in
higher-order logic.
Their key idea is to regard logical formulas as graphs by connecting a
propositional variable to its binder, while the other work such as
\citet{Irving_NIPS_2016} and \citet{Kaliszyk_arxiv_2017} deals with
them as token sequences.
This idea allows a DNN model to utilize structural information of
formulas and be invariant to names of bound variables.
%

\citet{DBLP:conf/lpar/LoosISK17} apply several off-the-shelf DNN
architectures to guide \emph{clause selection} of a saturation-based
first-order logic prover E~\cite{DBLP:conf/lpar/Schulz13}.
Given a conjecture to be proven, E generates a set of clauses from
logical formulas including the negated conjecture and investigates
whether a contradiction is derivable by processing the clauses one by
one; if a contradiction is found, the conjecture holds; otherwise, it
does not.
If E processes clauses that derive a contradiction early, the proof
search finishes in a small number of search steps.
Hence, clause selection is an important task in saturation-based
theorem provers including E.
Loos et al.\ use DNNs to rank clauses that are not processed yet and
succeed in accelerating the proof search by combining the DNN-guided
clause selection with existing heuristics.

This direction of enhancing the existing provers is orthogonal to our
present work. Although our goal is to generate proofs directly with
deep learning, rather than focusing on specific subproblems that are
important in theorem proving, we expect (as we discussed in
\sect{proof-synthesis}) that the combination of our approach with
these techniques is also beneficial to our technique.




\subsubsection{Formula proving}
Solving the Boolean satisfiability (SAT) problem by encoding problem
instances into neural networks has been attempted in early
days~\cite{Johnson_1989_JPDC}.
Recent work uses DNNs as a binary classifier of Boolean logical
formulas.
\citet{Bunz_2017_arxiv} represent a Boolean formula in conjunctive
normal form (CNF) as a graph where variable nodes are connected to
nodes that represent disjunctive clauses referring to the variables
and apply a graph neural
network~\cite{DBLP:journals/tnn/ScarselliGTHM09} to classify the
satisfiability of the formula.
Similarly NeuroSAT~\cite{Selsam_2018_arxiv} regards CNF formulas as
graphs, but it adopts a message passing model and can often (not
always) produce a Boolean assignment, which makes it possible to check
that the formula is truly satisfied.
\citet{Evans_2018_arxiv} tackles the entailment problem in the
propositional logic, that is, whether a propositional conjecture can
be proven under considered assumptions.
They also develop a new DNN model that classifies whether a given
entailment holds.
These lines of work do not guarantee the correctness of the solution.
Our work, although the procedure may not terminate, guarantees the
correctness of the returned proof.

\citet{DBLP:journals/corr/SekiyamaIS17} applied deep learning to proof
synthesis.
Their key idea is that the task of proof synthesis can be seen as a
translation task from propositions to proofs.
Based on this idea, they use a sequence-to-sequence
model~\cite{Sutskever_2014_NIPS}, which is widely used in machine
translation with deep learning, in order to translate a proposition to
its proof.
As shown in \sect{exp-proof}, our proposition-to-proof model
outperforms their model from the perspectives of (1) the number of
propositions that are successfully proved and (2) the time spent by
the proof-synthesis procedures.

\subsection{Neural program synthesis}
Synthesizing proofs from propositions can be regarded as synthesizing
programs from types via the Curry--Howard
isomorphism~\cite{Sorensen:2006:LCI:1197021}.
Program synthesis is one of the classical AI problems, and synthesis
with deep learning, dubbed \emph{neural program synthesis}, has been
studied recently.
A typical task of the neural program synthesis is to produce programs
satisfying given input-output
examples~\cite{DBLP:journals/corr/BalogGBNT16,DBLP:journals/corr/ParisottoMSLZK16,DBLP:conf/icml/DevlinUBSMK17,DBLP:conf/acl/YinN17}.
Although it appears to be difficult to transfer their approaches to
proof synthesis directly since the task of proof synthesis represents
a specification of a program by types (i.e., propositions), not
input--output pairs, there are similarities between them.
For example, the AST decoder based on the syntax of the target
language by \citet{DBLP:conf/acl/YinN17} is similar to that used in
our work in that we also construct an AST of a proof gradually,
whereas our proposition-to-proof model effectively uses the proof
obligations which do not appear in neural proof synthesis.
We expect that the ideas in neural program synthesis work in proof
synthesis as well to achieve better performance.

\subsection{Deep neural networks for tree structures}
Propositions and proofs have variable sizes, and a major way to handle
such variable-length data, especially, in natural language processing
is to deal with them as sequences.
However, such sequence representation collapses the structural
information contained in inputs.
%
Indeed, our work takes advantage of the fact that proofs can be
interpreted as derivation \emph{trees}, which makes it possible to
synthesize proofs gradually.
Besides propositions and proofs, many objects are
tree-structured---e.g., parse trees, hierarchical dependency graphs,
and index structures in databases---and recently there are many studies on tree
generation with DNNs.
\citet{Zhang_2016_NAACL} propose \emph{tree long short-term memory
  (TreeLSTM)} to construct tree structures.
TreeLSTM relates a parent and its children by a dependency path, which
connects a child node to the parent via the siblings.
This representation of node relationships needs more steps to pass
encoding features to a node.
As shown in \sect{exp-acc}, it would cause degradation of the
performance.
\citet{Dong_2016_ACL} generate tree-structured logical formulas from
natural sentences by a top-down decoder.
Their decoding method provides special nodes that link a parent to its
children, whereas our work does not need such nodes because we know
whether a node has children by looking at the inference rule of it.
\citet{Alvarez-Melis_2017_ICLR} also study a decoder for generation of
trees where each node has an arbitrary number of children.
Their decoder performs two predictions: one is whether a node has a
child; and the other is whether it has a sibling.
Unlike the task that they address, the number of children of an AST
node in the propositional logic is fixed and it is enough to predict
the kind of a node.
\citet{Mou_2016_AAAI} develop a tree-based convolutional neural
network (TBCNN), which calculates a feature vector of a node by using
vectors of nodes near it.
While it is similar to the AST convolution and the aggregation layer
in our work, there is a difference for each.
First, the AST convolution refers to all adjacent nodes including the
parent, whereas the TBCNN considers only children.
Second, the aggregation layer can be seen as ``fold'' on trees with
feature vectors and the produced single vector should contain features
of all nodes in a tree.
The TBCNN uses a max-pool to integrate feature vectors of nodes into a
single vector, that is, it produces a vector each dimension of which
has the value maximum among the corresponding dimensions of feature
vectors of the nodes.
Although max-pools are commonly used in usual (not tree-based)
convolutional neural networks~\cite{Alex_2012_NIPS}, it is unclear
that gathering only maximum values does not drop any important feature
of nodes in a tree.
%

\section{Conclusion}
\label{sec:conc}
%
We present an approach to applying deep learning to the APS problem.
We statistically formulate the APS problem in terms of probabilities so that
we can quantify the likelihood of a term being a correct proof of a proposition.
From this formulation, we show that this likelihood can be calculated by using the likelihood of an inference rule being applied at a specified position in a proof, which enables us to synthesize proofs
gradually.
To approximate this likelihood, we develop a DNN that we call a proposition-to-proof model.
Our DNN model encodes the tree representation of a proposition and decodes it to
estimate an inference rule to be applied by using the proof obligation to be discharged effectively.
We train the proposition-to-proof model on a dataset of automatically
generated proposition-proof pairs and confirmed that the trained model
achieves 96.79\% accuracy in the inference-rule estimation, though
there is still room for improvement.
We also develop a proof synthesis procedure with the trained DNN model and
show that it can synthesize many proofs of a proposition in short
time.

Our exploration of APS along with deep learning is still at the early
stage; there are many challenging tasks to be addressed.
One of the important challenges is to extend the target logic to more
expressive ones such as first-order logic and higher-order logic.
For example, first-order logic introduces the notions of predicates
and quantification.
To learn a feature representation of a predicate, we may need a DNN
model that takes the ``meaning'' of a predicate into account.
Quantification not only makes formulas complicated but also requires
us to deal with the problem of instantiation.
%
Another important notion that we need to deal with is the induction
principle.
%
%
With the extension of the logic, it is expected that a problem with
datasets happen.
One promising way to address it is, as done by
\citet{Kaliszyk_arxiv_2017}, making a dataset from publicly available
proofs.
Furthermore, the creation of a benchmark collecting challenging tasks
related to APS is crucial for the development of APS with deep
learning, as ImageNet~\cite{ILSVRC15} contributes to the advance of
image processing.

Another future direction is improvement of a model.
Our model is expected to have access to assumptions via the feature
vector of a given proposition.
However, it may be more useful to encode a set of assumptions
directly, as we encoded proof obligations in this work.
A problem with it is that the number of assumptions is not fixed; DNNs
are good at handling objects with a fixed size but require efforts to
deal with variable-sized data.
Another possible issue is the vanishing gradient problem; gradients in
very deep neural networks often vanish, which makes learning
difficult.
Since our tree-structured model can be considered to have
variable-length nonlinear layers and become deeper as propositions
and/or proofs are larger, that problem would be more serious when we
deal with larger propositions and proofs than the present work.
We expect that the recent progress in research to address this problem
works well also in our settings; especially, residual
blocks~\cite{he2016deep} and LSTMs~\cite{LSTM} are promising
workarounds.


\begin{acks}                            
  This work is partially supported by JST PRESTO Grant Number JPMJPR15E5, Japan.
\end{acks}

\bibliography{main}



\end{document}